\documentclass[10pt,twocolumn,letterpaper]{article}

\usepackage{iccv}
\usepackage{times}
\usepackage{epsfig}
\usepackage{graphicx}
\usepackage{amsmath}
\usepackage{amssymb}

\usepackage{algorithm,algorithmic}
\usepackage{amsmath}
\usepackage{amsfonts} 
\usepackage{booktabs}
\usepackage{url}
\usepackage{cite}
\usepackage{microtype}
\usepackage{multirow}
\usepackage{subcaption}
\usepackage[accsupp]{axessibility}
\usepackage[pagebackref=true,breaklinks=true,letterpaper=true,colorlinks,bookmarks=false]{hyperref}

\iccvfinalcopy 

\def\iccvPaperID{5894} 

\ificcvfinal\fi

\begin{document}

\title{Learning to Cut by Watching Movies}

\author{
Alejandro Pardo$^{1}$ \quad
Fabian Caba Heilbron$^{2}$ \quad
Juan León Alcázar $^{1}$ \quad
Ali Thabet $^{1}$ \quad
Bernard Ghanem$^{1}$ \\
$^{1}$King Abdullah University of Science and Technology (KAUST) \quad $^{2}$Adobe Research \\
{\tt\small\{alejandro.pardo,juancarlo.alcazar,ali.thabet,bernard.ghanem\}@kaust.edu.sa \quad caba@adobe.com} \\
{\textit{\textbf{\url{https://alejandropardo.net/publication/learning-to-cut/}}}}
}
\maketitle
\ificcvfinal\thispagestyle{empty}\fi

\begin{abstract}
  Video content creation keeps growing at an incredible pace; yet, creating engaging stories remains challenging and requires non-trivial video editing expertise. Many video editing components are astonishingly hard to automate primarily due to the lack of raw video materials. This paper focuses on a new task for computational video editing, namely the task of raking cut plausibility. Our key idea is to leverage content that has already been edited to learn fine-grained audiovisual patterns that trigger cuts. To do this, we first collected a data source of more than $10K$ videos, from which we extract more than $255K$ cuts. We devise a model that learns to discriminate between real and artificial cuts via contrastive learning. We set up a new task and a set of baselines to benchmark video cut generation. We observe that our proposed model outperforms the baselines by large margins. To demonstrate our model in real-world applications, we conduct human studies in a collection of unedited videos. The results show that our model does a better job at cutting than random and alternative baselines.
\end{abstract}


\section{Introduction}
\label{sec:introduction}

\begin{figure}[ht!]
    \begin{center}
        \includegraphics[width=0.95\linewidth]{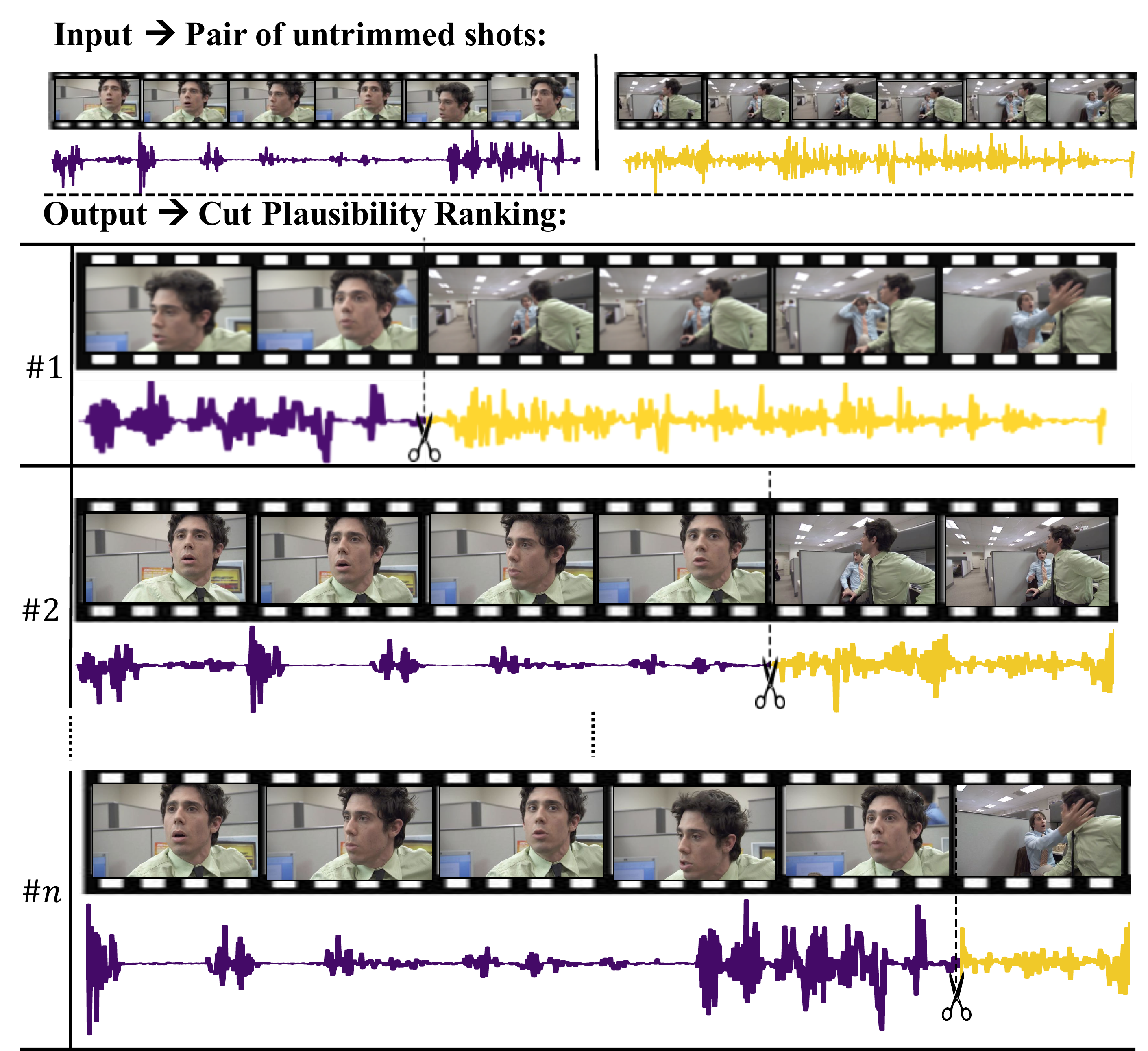}
    \end{center}
    \caption{\textbf{Ranking Cut Plausibility.} We illustrate the process of ranking video cuts. The first row show a pair of untrimmed videos (raw footage) as the input. The output would be the ranking of all the possible cuts across the pair of shots. Ideally, the top ranked cuts should be the more plausible cuts providing a smooth transition between the shots, and the worst cuts would be places in where there is break of spatial-temporal continuity. 
    }
    \label{fig:teaser}
\end{figure}

The lack of video editing expertise is a common blocker for aspiring video creators. It takes many training hours and extensive manual work to edit videos that convey engaging stories. Arguably, the most time-consuming and critical task in video editing is to compose the right \emph{cuts}, \ie, decide how (and when) to join two untrimmed videos to create a single clip that respects continuity editing\cite{smith2006attentional}. To the untrained eye, cutting might seem easy; however, experienced editors spend hours selecting the best frames for cutting and joining clips. In light of this complexity, it is pertinent to ask: could artificial systems rank video cuts by how plausible they are?

Before delving into addressing the question above, it is worth defining the task of video cut ranking in detail. As Figure \ref{fig:teaser} illustrates, given two untrimmed input videos, the goal is to find the best moments (in each video) to trigger cuts, which join the pair into a single continuous sequence. A key challenge is to generate videos that make the audience believe actions unfold continuously. This type of cutting is often called continuity editing and aims to evoke an illusion of reality \cite{smith2006attentional, filmgrammar}, even though, the source videos could be recorded at different times. Figure \ref{fig:teaser} shows a typical trigger for cuts -- the moment when the speaker changes. In practice, the director could give the shot order via a storyboard or script, and it is the editor's job to realize which patterns make smooth transitions between shots. Our hypothesis is that many of those cut-trigger patterns can be found by carefully analyzing of audio-visual cues.

Despite its importance, potential impact, and research challenges, the computer vision community has overlooked the video cut ranking problem. While there has been significant progress in shot boundary detection \cite{pal2015video}, video scene segmentation \cite{rao2020local}, video summarization \cite{money2008video}, and video-story understanding \cite{bain2020condensed, huang2020movienet}, few to no works have focused on pushing the envelope of computational video editing. 

The most relevant works at addressing video cut ranking and generation are found in the graphics and HCI communities \cite{berthouzoz2012tools, truong2016quickcut, leake2017computational, wang2019write, fried2019text}. These attempts take on a different perspective and focus on human-computer experiences for faster video editing. Yet, they still demand extensive work from an editor in the loop. We hypothesize that the cut ranking task has been neglected due to the lack of data, \ie, raw footage, and its corresponding cuts done by an editor. 

In this paper, we introduce the first learning-based method to rank the plausibility of video cuts. It is important to note that we do not have access to the raw footage for each shot since it is difficult, \ie, requires expertise and extensive manual work, to gather a dataset of raw videos with the corresponding edits and cuts. Therefore, Our key idea is to leverage \textit{edited video content} to learn the audio-visual patterns that commonly trigger cuts. 

While this data is not the same as the real-world input for generating and ranking cuts, we can still model the audio-visual data before and after the cut, thus modeling what good cuts look like. Additionally, this type of data can be found abundantly, which enables the development of data-driven models. Our results show that a model learned to solve this \textit{proxy task} can be leveraged to practical use cases.
 
Our approach begins with a pair of consecutive shots that form a cut (similar to the bottom row in Figure \ref{fig:teaser}). We look for a representation that discriminates between the good cuts (actual cuts found on edited video) against all alternative options (random alignments). To achieve this goal, we first collect a large-scale set of professionally edited movies, from which we extract shot boundaries to create more than 260K cuts and shot pairs. Using this new dataset, we train an audio-visual model, \emph{Learning-to-Cut}, which learns to rank cuts via contrastive learning. Our experimental results show that, while challenging, it is possible to build data-driven models to rank the plausibility of video cuts, improving upon random chance and other standard audio-visual baselines.

\noindent\textbf{Contributions.} To the best of our knowledge, we are the first to address video cut ranking from a learning-based perspective. To this end, our work brings two contributions. 

\noindent\textbf{(1)} We propose Learning-to-Cut, an audio-visual approach based on contrastive learning. Our method learns cross-shot patterns that trigger cuts in edited videos (Section \ref{sec:method}).\\
\noindent\textbf{(2)} We introduce a benchmark and performance metrics for video cut ranking, where we show the effectiveness of Learning to Cut. Moreover, we showcase that expert editors more likely prefer the cuts generated by our method as compared to cuts randomly ranked and other baselines (Section \ref{sec:experiments}).
\section{Related Work}
\label{sec:related}
\noindent\textbf{Computational Video Editing.} Earlier research in computational video editing focuses on designing new experiences that speed up the video creation process \cite{berthouzoz2012tools, truong2016quickcut, leake2017computational, wang2019write, fried2019text}. For instance, Leake \etal propose a system for the automatic editing of dialogue-driven scenes \cite{leake2017computational}. This system takes as input raw footage, including multiple takes of the same scene, and transcribed dialogue, to create a sequence that satisfies a user-specified film idiom \cite{filmgrammar}. Although this method offers a modern video editing experience, it still relies on a rule-based mechanism to generate the cuts. Another line of work focuses on designing transcript-based video editing systems. To cite an example, QuickCut \cite{truong2016quickcut} and Write-A-Video \cite{wang2019write} develop user interfaces that allow aspiring editors to create video montages using text narrations as input. Although significant progress has been made to create better video editing experiences, it is still an open question of whether learning-based approaches can advance computational video editing. Our work provides a step towards that direction by introducing a benchmark for ranking the plausibility of video cuts and an audio-visual method that learns how to approximate them, without fixed rules and trained from an unconstrained set of professionally edited videos.

\noindent\textbf{Long-term video analysis.} Many efforts have been made to develop deep learning models that analyze and understand long-term information \cite{yu2020relationship, gupta2018linear} and model relationships in long video formats \cite{bain2020condensed,huang2020movienet, gu2018ava}. Recently, Bain~\etal \cite{bain2020condensed} collected a dataset that contains movie scenes along with their captions, characters, subtitles, and face-tracks. Similarly, Huang~\etal \cite{huang2020movienet} created a dataset that comprises complete movies along with their trailers, pictures, synopses, transcripts, subtitles, and general metadata. Based on these datasets, the research community has developed solutions for new movie-related tasks, such as: shot-type classification \cite{rao2020unified}, movie-scene segmentation \cite{rao2020local}, character re-identification and recognition \cite{huang2018person, xia2020online, huang2020caption, brown2020playing}, trailer and synopsis analysis \cite{Xiong_2019_ICCV, huang2018trailers}, and visual-question answering in movies \cite{jasani2019we, garcia2020knowledge}. Unlike previous works in long-term video analysis, our work centers around the video creation process. Another line of work worth a bit more closer to ours is video summarization \cite{tvsum,vs_gong2014diverse,vs_otani2019rethinking}. These approaches are typically given \textit{one long video stream} and their task is to select and shorten shots while keeping the semantic meaning of the composed video. Although video summarization techniques compose shot sequences, they tend to disregard critical aspects of video editing such as maintaining spatial-temporal continuity across shots. To the best of our knowledge, our work is the first benchmark studying the plausibility of cuts. As there are limited previous works aligning with our task, we define a set of initial baselines, and a novel approach for evaluating the ranking quality of video cuts.

\noindent\textbf{Cross-Modal Representations.} The joint exploration of multiple modalities is a hot topic in the research community. Several works have explored self-supervised learning to learn semantic embeddings using cross-modality supervision by using text, audio, and visual streams \cite{alwassel2020selfsupervised, alayrac2020self, patrick2020multi, arandjelovic2017look, aytar2017see, korbar2018cooperative, owens2016ambient, owens2018audio, tian2019contrastive, miech2020end, kalantidis2020hard}. Moreover, recent works have used already pre-trained embeddings from different modalities on video retrieval tasks \cite{gabeur2020multi, liu2019use, yu2020video, escorcia2019temporal, lei2020tvr, li2020hero, amrani2020noise}, active speaker detection \cite{chung2016out, chung2018voxceleb2, alcazar2020active}, and sign spotting \cite{momeni2020watch}. Our method adopts multimodal streams and backbones to model audio-visual signals and learn the patterns that commonly trigger a cut.

\noindent\textbf{Contrastive Losses for Video.} Rapid progress has been attained in video understanding research by leveraging contrastive losses \cite{gutmann2010noise, miech2020end}. The InfoNCE loss \cite{gutmann2010noise}, particularly, has gained tremendous popularity with the advent of self supervised learning \cite{tschannen2020self, gordon2020watching, qian2020spatiotemporal, wang2020selfsupervised, alayrac2020self, morgadoNIPS20, han2019video, han2020self}. The idea behind this loss is simple; given a query representation, its goal is to maximize the similarity with a corresponding positive sample, but minimize the similarity concerning a bag of negatives. In the video domain, the InfoNCE loss and its variants (\eg, MIL-NCE \cite{miech2020end}) have been used for learning: general video representations \cite{tschannen2020self, qian2020spatiotemporal, gordon2020watching}, joint text-video embeddings \cite{miech2020end, alayrac2020self}, or audio-visual models \cite{morgadoNIPS20, alayrac2020self, afouras2020selfsupervised}. Following the widespread adoption in the video community, our work leverage the InfoNCE loss \cite{gutmann2010noise} to learn a representation that encodes how good video cuts look (and sound) compared to all the other cut alternative for a pair of shots.
\section{Learning to Cut}\label{sec:method}

\subsection{Leveraging Edited Video}
\label{sec:benchmark}

As a reminder, a shot is a continuous take from the same camera, and a cut occurs between a pair of shots (section \ref{sec:introduction}). We introduce a data source devised for the task of learning suitable video cuts from professionally edited movies. The primary purpose of this data collection is to leverage already edited content for learning fine-grained audio-visual patterns that trigger cuts in the video editing process.  It is composed of $10,707$ movie scenes along with $257,064$ cuts. This dataset includes several streams of data for every shot, including visual, audio, and transcripts. Around $32\%$ of shots in the dataset include speech; the remaining shots without speech come mainly from action-driven scenes. In the supplementary material we provide additional statistics.

\begin{figure}[t!]

\begin{subfigure}{.49\linewidth}
  \includegraphics[width=\linewidth]{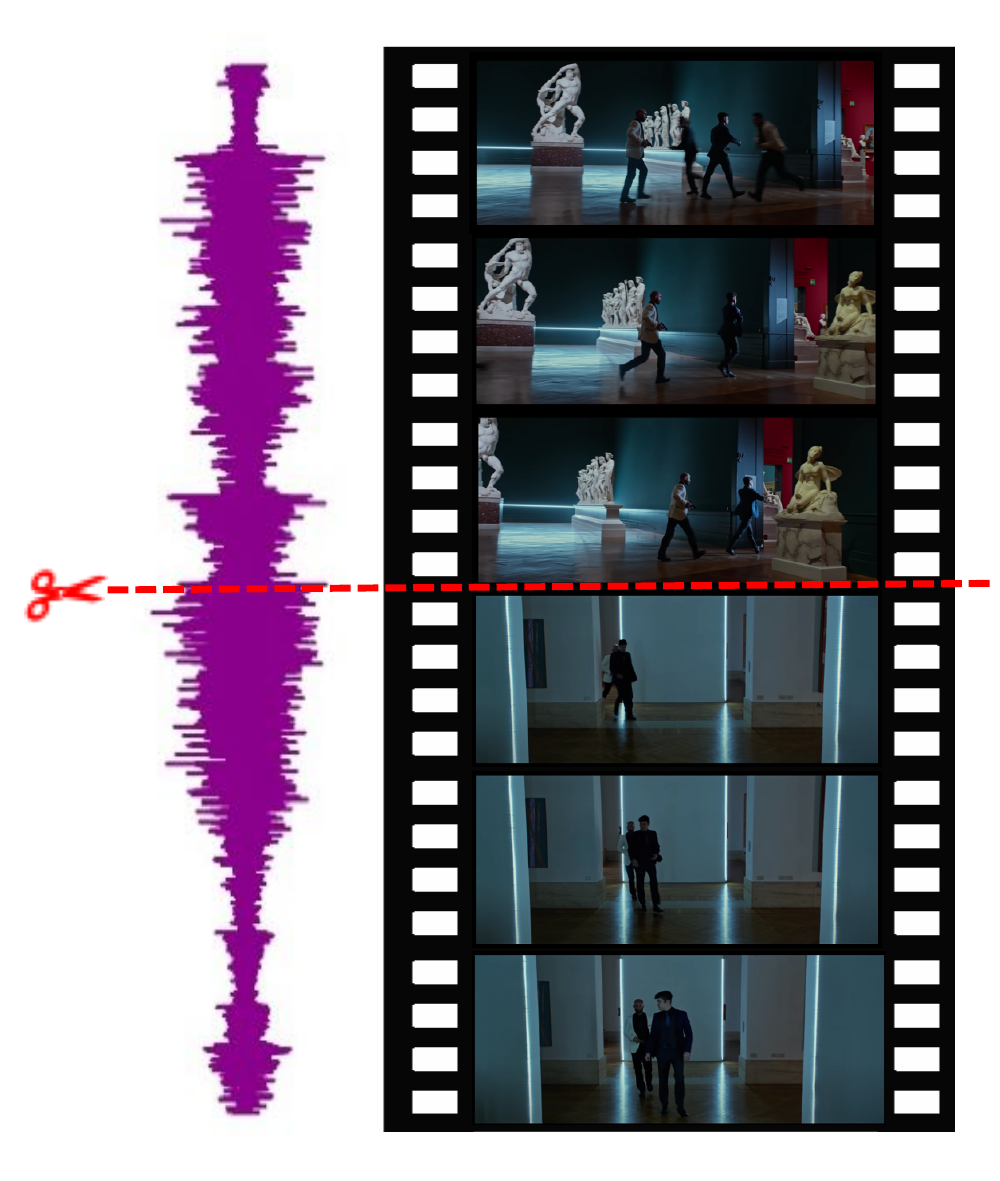}
  \caption{}\label{fig:cut_on_action}
\end{subfigure}
\
\begin{subfigure}{.49\linewidth}
  \includegraphics[width=\linewidth]{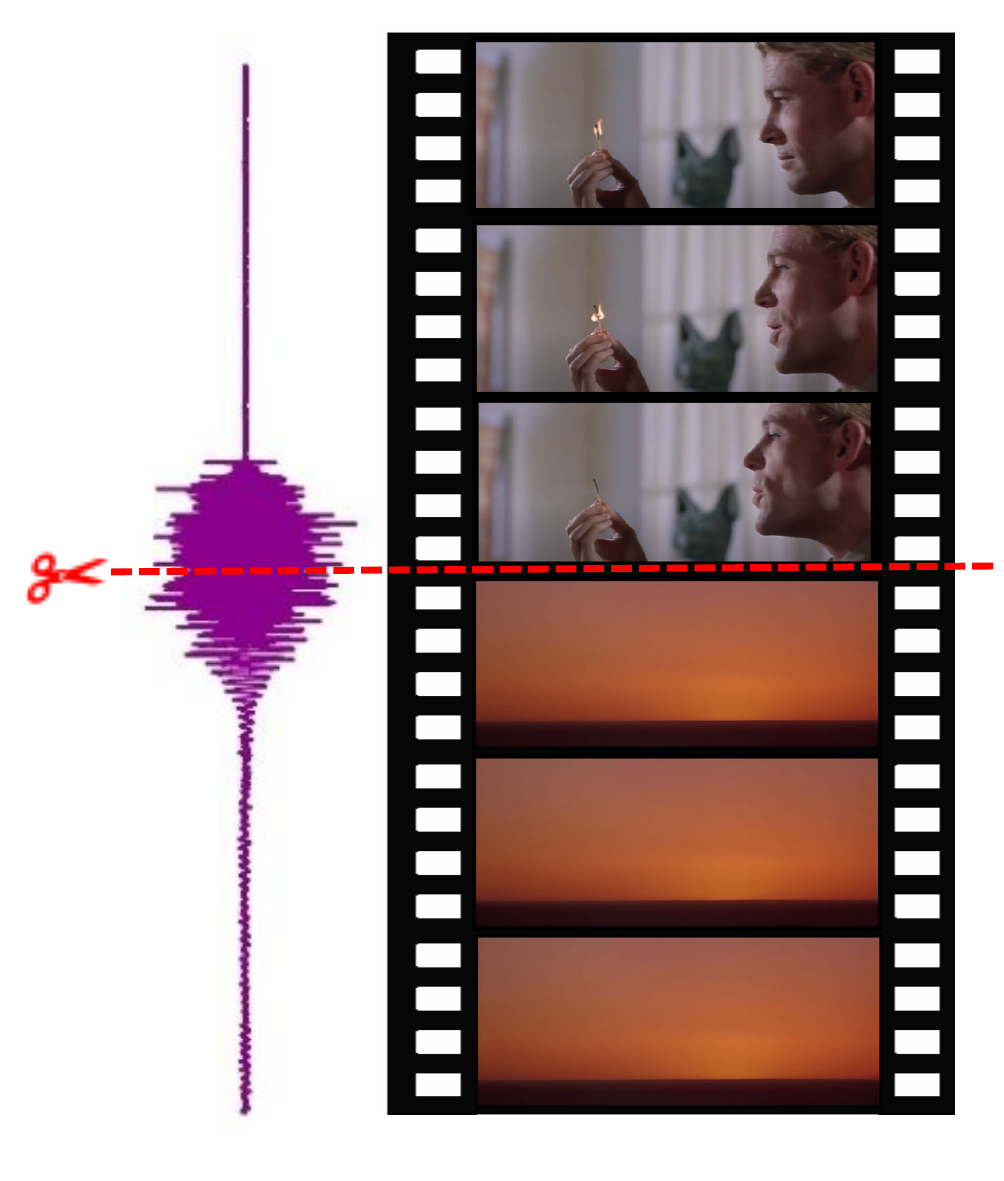}
  \caption{}\label{fig:sound_match_cut}
\end{subfigure}
\\
\begin{subfigure}{.49\linewidth}
  \includegraphics[width=\linewidth]{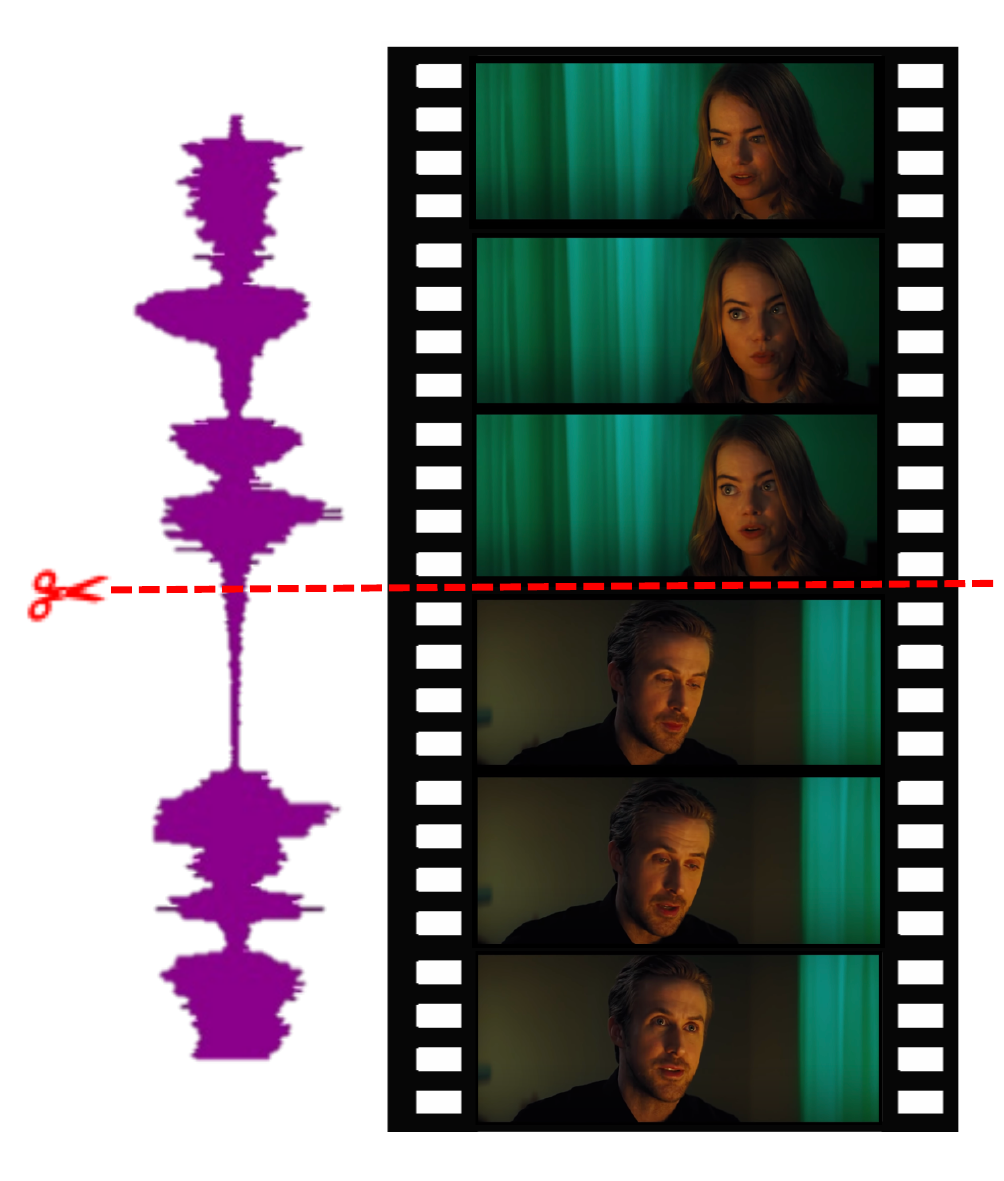}
  \caption{}\label{fig:speaker_change_cut}
\end{subfigure}
\
\begin{subfigure}{.49\linewidth}
  \includegraphics[width=\linewidth]{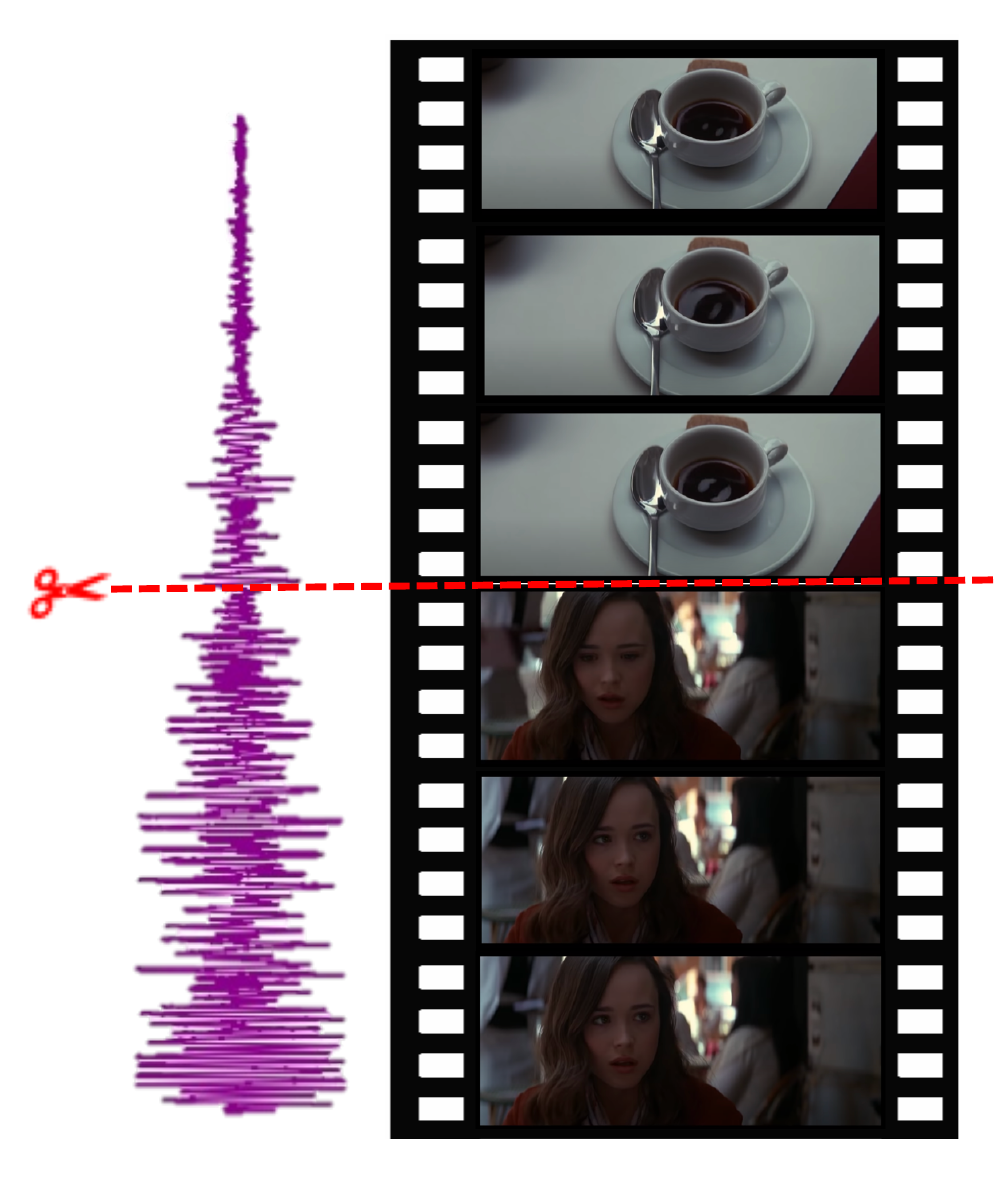}
  \caption{} \label{fig:reaction_cut}
 \end{subfigure}

\caption{\textbf{Examples of the retrieved content.} Figure \ref{fig:cut_on_action} shows a cut driven by the visual action of entering through a door. Figure \ref{fig:sound_match_cut} shows a cut driven by the matching audio between two sounds. Figure \ref{fig:speaker_change_cut} shows a cut driven by a conversation an its audio-visual signal. Finally, \ref{fig:reaction_cut} shows a visual-reaction cut that is driven by an audio-visual action.}
\label{fig:dataset-examples}
\end{figure}

\noindent\textbf{Gathering edited video.} Movies are a great source of video data containing creatively edited video. Following Bain et al. \cite{bain2020condensed}, we downloaded $10,707$ videos from \textit{MovieClips} \footnote{\href{https://www.youtube.com/c/MOVIECLIPS/videos}{MovieClips: Source of the videos}}. Each of these videos correspond to a single movie scene (a section of the movie occurring in a single location with a continuous and condensed plot). Then, we automatically detected shot boundaries using \cite{gygli2017ridiculously}. To asses the quality of the detections, we verified $5,000$ of them and found an error rate of $4.34\%$ (217 errors). Typical errors include shots with heavy visual effects, and partial dissolves. 

We find that each scene contains $114$ shots on average. Thus, an editor has to make more than a hundred cuts for every scene in a movie. Such a level of involvement further shows the complexity and time-consuming nature of the editing process. An appealing property of the chosen video source, MovieClips, is that it has weekly uploads of famous movie scenes, along with their metadata. Since our shot annotations are automatically generated, we can easily augment the dataset in the future. 

\noindent\textbf{Data source samples.} Figure \ref{fig:dataset-examples} shows examples from the collected data. As specified before, the cut can be driven by visual, audio, and audio-visual cues. In Figure \ref{fig:cut_on_action}, the visual action of entering the door triggers the cut, while the audio stream does it in Figure \ref{fig:sound_match_cut}. In the latter, the editor matched two similar sounds from two different space-time locations; this example shows a visually discontinuous cut, where the audio matches perfectly between the two shots. The last two examples are cuts driven by audio-visual cues.

On one hand, Figure \ref{fig:speaker_change_cut} is triggered by the nature of a fluid conversation -- we can appreciate that the cut happens after the active speaker changes. On the other hand, the cut in Figure \ref{fig:reaction_cut} is driven by an action observed in both streams -- the scene shows a person's reaction to an audio-visual action. Note that these are just some of the triggers for cuts, and many others exist, making it hard to list and model each of them independently. Therefore, our next goal is to leverage this data to learn cut triggers in a data-driven fashion. 

\noindent\textbf{Dataset.} We use edited movies to learn the proxy task of cutting edited videos. To do so, we split our dataset into train and validation sets. Since our task requires both positive and negative examples per pair of shots, we ignore shots shorter than one second for both sets. We use the frame-cut info to remove all the snippets that contain shot transitions within them to avoid the degenerate shortcut of learning how to detect shots. After these two filters, the training set consists of $7,494$ scenes with $177,987$ shots, and the validation set consists of $3,213$ scenes with $79,077$ shots.

\subsection{Proxy task}
We propose a proxy task closely related to the actual editing process of cutting (and merging) two shots. We define a \textit{snippet} as a short time range within a shot. A snippet is fully contained within a shot and hence is much shorter. 
Our proxy task consists of learning which pair of snippets (from a set composed of all clips from consecutive shots) is the best to stitch together.
Since we know that the editing process is mainly done in sequential order (stitching the left shot to the right shot), we can try to solve a local (two-neighboring-shots) proxy task by recurrently asking the following question: which snippets in both of the shot videos are the best to stitch together?.  The best place to cut in each of the two shots are these two retrieved snippets.

Our proxy task resembles the actual editing process, where the editor needs to pick a place to cut based on what they want to show in the next shot. However, it is not exactly the same, since we do not have the part of the shots that are cut out during the editing process. Obtaining such data is challenging as it requires video editing expertise to select the appropriate cuts, and at the same time, it is hard to find large-scale footage without edits. Although not all possible clips are available during training, we argue (and show with experiments) that our proxy task, paired with a vast number of edited videos, provide enough supervisory signals to learn the audio-visual cues and patterns that trigger the cuts.

We can address the proxy task in several ways; however, there is one that best fits the task editors solve in continuity editing \cite{smith2006attentional}: \emph{maximize the smoothness of the transition or minimize the discontinuity in the semantics of the scene across a cut}. Furthermore, given the task's artistic nature, there is not necessarily a single correct answer. There might be more than one place in the videos where a cut can be placed and several pairs that would make semantic sense. Considering the previous observations, we decide to model this task as a ranking problem. We also build a small temporal window near the actual shot boundaries, and consider all clips within this window as appropriate cutting points. As a result we don't aim at retrieving only the clips at the very end/beginning of adjacent shots, but we also consider as valid some highly similar clips with significant temporal overlap. We approach this task by using Contrastive Learning. In fact, we aim to find a space, where the representations of clip pairs that belong together are close to each other and far from all others.

\noindent\textbf{Technical description.} Given a ground-truth cut formed by a pair of shots $(S_L, S_R)$, we aim to find the best pair of snippets to stitch together. We define $S_L$ as the set  $\{a_i \rvert i \in \mathbb{N}, 0 \leq i \leq t_L \}$, and $S_R$ as $\{b_j \rvert j \in \mathbb{N}, 0 \leq j \leq t_R \}$, with $t_L$ and $t_R$ being the number of snippets contained in the shots, respectively. Our proxy task consists of ranking the set of all pairs of snippets $\{(a_i, b_j) \rvert i,j \in \mathbb{N}, 0 \leq i \leq t_L, 0 \leq j \leq t_R\}$. We know that the pair that should appear on top of the ranking is the one formed by the temporally adjacent snippets $(a_{t_L}, b_0)$. The rest of the pairs 
are negative samples that we want to avoid retrieving.

\noindent\textbf{Potential short cuts.} One may think that this artificial task can be easily solved by any machine learning model by learning shortcuts and apparent biases, for instance, always ranking first the pair composed by the last clip of the left-hand-side shot and the first of the right-hand-side shot. Additionally, if we include the clip in which the actual shot transition occur in our training data, we would end up learning a shot detector. We avoid these two degenerate cases by not including any information of the temporal position of a clip within the shot and also ignoring all the clips that contain the transition from one shot to another. 



\subsection{Learning to Cut Model}

\begin{figure}[t!]

  \hspace*{10pt}\includegraphics[width=\linewidth]{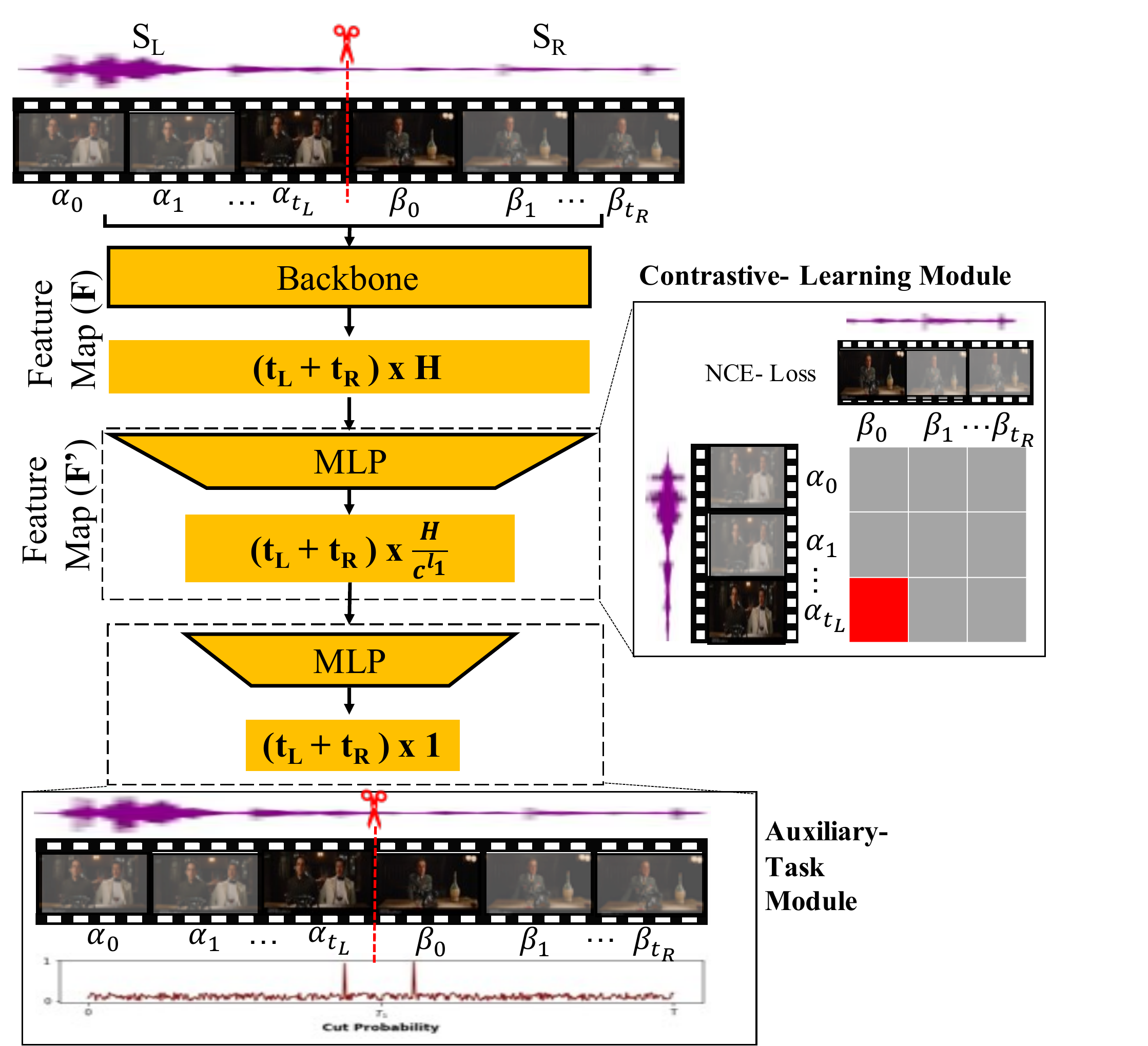}  
  \caption{\textbf{Learning to Cut Model}. Our model takes as input the audio-visual streams of a pair of shots. It computes their snippet pair-wise similarity and backpropagates it using the NCE-loss. Finally, it predicts a score per snippet that indicates whether or not it is a good place to cut. In the figure the non-blurred frames represent positive pairs. The NCE-loss only positive sample is represented with red on the grid, which corresponds to the neighbour friends of the left and right shots. 
  }
\label{fig:pipeline}
\end{figure}

Our cut ranking model consists of three main modules: a Feature Extraction step, a Contrastive Learning module, and an Auxiliary Task module. We first extract audio-visual features from candidate snippets. Then, the Contrastive Learning Module refines these features such that pairs of clips forming good cuts yield higher similarity score than others. The last module consists of an auxiliary task that guides the model by predicting whether individual snippets are good places to cut or not. Our pipeline is illustrated in Figure \ref{fig:pipeline}. Below, we describe each module in detail.

\subsubsection{Feature Extraction} This task is inherently multimodal, since editors use fine-grained visual and audio signals from the shots to determine the cuts. Hence, we use both audio and visual backbones. It has been shown that effective visual features for video analysis come from short-term spatio-temporal architectures \cite{ji20123d, taylor2010convolutional, tran2015learning, carreira2017quo, wang2018temporal, feichtenhofer2019slowfast}. Similarly, effective audio architectures represent each time instance with a temporal window of fixed size and its corresponding spectogram \cite{takahashi2016deep, hershey2017cnn, Chen20}.  We extract features for each snippet of $T$ frames with temporal stride of $\Delta$. We input each one of the shots $S_{\{LR\}}$ with $N_{\{LR\}}$ frames to extract $t_{{\{LR\}}}=Floor(\frac{N_{\{LR\}}-T}{\Delta})+1$ snippets, and extract shot-level features $F_{\{L,R\}}$ by concatenating each of its snippet-level feature maps of dimension $H$. Formally $F_L=[\alpha_1;...;\alpha_{t_L}]$ and $F_R=[\beta_1;...;\beta_{t_R}]$ where $\alpha,\beta \in \mathbb{R}^{1 \times H}$.

\subsubsection{Contrastive Learning Module} 
This module consists of a Multi-Layer Perceptron (MLP) with $l_1$ layers interleaved with a ReLU activation. It receives the shot-level features $F_{\{LR\}}$. Each layer reduces the size of the previous layer by $c$, similar to the projection head used in \cite{chen2020simple}. It produces a feature $F^{\prime}_{\{LR\}}$ where each of its elements belongs to $\mathbb{R}^{1 \times \frac{H}{c^{m_1}}}$. We compute each of the possible snippet pairs between $F^{\prime}_L$ and $F^{\prime}_R$, and we produce the annotations as follows: the positive sample is the pair $(\alpha_{t_L}, \beta_0)$, while the negative samples are the set $N(\alpha,\beta) = \{(\alpha_{i^\prime}, \beta_{j^\prime}) \rvert i^\prime,j^\prime \in \mathbb{N}, 0 \leq i^\prime \leq t_L-1, 1 \leq j^\prime \leq t_R \}$. We aim at bringing the features into a different space, in which snippet pairs that are good to stitch together are close to each other and far from the rest. To enable this, we use a Noise-Contrastive-Estimation (NCE) loss \cite{gutmann2010noise} defined as:

\begin{equation}
  NCE(S)= -\log \left( \frac{e^{  \alpha_{t}^{\top} \beta_{0} }}
                        {e^{ \alpha_{t_L}^{\top} \beta_{0} } +  \sum\limits_{N(\alpha,\beta)} e^{  \alpha_{i \prime}^{\top} \beta_{j \prime} }}  \right)
\end{equation}

This loss encourages that the positive pair (numerator) attains the highest correlation score among all pairs (denominator). We expect it to align well with the ranking task we are interested in. By maximizing the similarity between the two adjacent snippets from each pair of shots, we expect the model to learn smooth transitions between shots, thus mimicking what editors do in continuity editing \cite{smith2006attentional}. 

\subsubsection{Auxiliary Task Module} 
It consists of an MLP with $l_2$ FC-layers to produce a single value per snippet, which is then passed through a sigmoid function. 
This module receives the feature map $F^{\prime}_{\{LR\}}$ and produces a vector $v_{\{LR\}} \in \mathbb{R}^{(t_L+t_R) \times 1}$, one score per snippet. The higher the score, the higher the probability $p_{y_i}$ for a snippet to be a good place to cut. Remember that our auxiliary task aims to answer whether or not each snippet is a good place to cut. We believe that this is a reasonable task to guide the learning process, since there are some pre-established rules in video editing that drive the cutting process \cite{filmgrammar, berthouzoz2012tools}. And so, we hypothesize that individual snippets can contain some information about these cues and can guide the model to cut more precisely. We propose to learn this auxiliary task by classifying each of the snippets as a good place to cut or not. In this case, the only positive snippet is $\alpha_{t_L}$. We iterate over each snippet in $S_L$ to calculate the Binary Cross-Entropy (BCE) Loss per shot:

\begin{equation}
\small
    BCE(S)=- \frac{1}{t} \sum_{i}^{t} y_i \log(p_{y_i}) + (1-y_i)  \log(1-p_{y_i})
\end{equation}

\noindent Therefore, our model optimizes for both two tasks jointly and the losses are combined as follows:
\begin{align} 
    Loss(S) = \lambda_1 \cdot NCE(S) + \lambda_2 \cdot BCE(S) \label{equation:loss}
\end{align}
\section{Experiments}
\label{sec:experiments}

This section describes the experimental pipeline that we follow to validate our approach's effectiveness in learning  to rank cuts. We detail our performance metrics, implementation details of our model, and introduce baseline approaches. Then, we then study our method's performance on our dataset \ref{sec:benchmark} by comparing to the baselines and ablating each of its components. Finally, we use our model to rank cuts in an unedited set of video footage and assess the quality of the top ranked results via human studies.

\subsection{Experimental Setup}

\noindent\textbf{Metrics.} We aim to define metrics that measure the quality of automatically ranked video cuts. One desirable property of automatic cut generation method is the ability to rank the good cuts with higher confidence across a test subset. To capture this property, we measure Recall at $(\eta K)$, ($R@\eta K$), where $\eta \in \{1,5,10\}$ is a value that controls the number of retrieved results and $K$ is fixed to the number of ground truth cuts in the entire inference set. In our experiments, $K \approx 80000$, which corresponds to the number of shot pairs or cuts in the validation subset. To account for the ambiguity of cuts, we measure $R@(\eta K)$ using different temporal precision ranges when counting the true positive cuts. Our intuition is that cuts near the ground-truth might be equally good and still useful for many practical scenarios. Therefore, we introduce a distance to cut $d$, which augments the number of positives with all clip pairs that are within $d$ seconds from the ground-truth cut. In practice, $d$ is the number of clips away from the ground truth cut. We report $R@(\eta K)$ for three distance values $d\in\{1,2,3\}$. These metrics allow us to study the performance of methods under different use cases. For instance, if someone plans to evaluate performance for automatic retrieval, $R@1K$ would be the best fit; however, if there will be a human-in-the-loop, $R@10K$ seems a more appealing metric to optimize, as a human could select the best cut from a small list of candidates. Conversely, measuring performance at $d=1$ and $d=3$ resemble cases where an application requires high-precision (\eg editing for professional film) and low-precision cuts (\eg editing for social media), respectively.

\begin{table*}[!ht]
\centering
\footnotesize
\begin{tabular}{@{}c@{\hspace{0.4em}} l c@{\hspace{0.2em}} ccc c@{\hspace{0.2em}} ccc c@{\hspace{0.2em}} ccc c@{\hspace{0.2em}} c@{}}
\toprule
&& \phantom{} & \multicolumn{3}{c}{$d=1$} &  \phantom{} & \multicolumn{3}{c}{$d=2$} &  \phantom{} & \multicolumn{3}{c}{$d=3$} & \phantom{} \\
\cmidrule{4-6} \cmidrule{8-10} \cmidrule{12-14} 
& Model && R$@1K$ & R$@5K$  & R$@10K$ && R$@1K$ & R$@5K$  & R$@10K$  && R$@1K$ & R$@5K$  & R$@10K$  \\
\midrule
&Random && 0.60 & 3.25 & 6.56 && 1.69 & 9.44 & 18.45 && 3.55 & 17.62 & 33.75\\
\midrule
&Visual Raw Features && 1.11 & 2.78 & 5.17 && 2.03 & 5.35 & 10.26 && 2.74 & 7.46 & 14.73\\
&Audio-visual Raw Features && 1.17 & 6.37 & 11.73 && 2.51 & 13.15 & 24.25 && 3.73 & 19.33 & 34.97\\
\midrule
&Learning to Cut (Single-stage) && 2.89 & 9.82 & 16.15 && 5.37 & 18.24 & 30.01 && 7.10 & 24.18 & 40.21 \\
&\textbf{Learning to Cut (Full)} && 8.18 & 24.44 & 30.40 && 15.30 & 48.26 & 59.50 && 19.18 & 64.30 & 79.42 \\
\bottomrule
\end{tabular}
\caption{\textbf{Comparison with baselines.} We show the different baselines compare to our method, with and without two-stage inference. We observe that our method beats all of the baselines by large margins in all the metrics. Also, the two-stage inference significantly improves upon its one-stage counterpart. We report $R@\{1,5,10\}$ with $d\in\{1,2,3\}$.} 
\label{table:baselines_comparison}
\end{table*}

\begin{table*}[h!!]
\centering
\footnotesize
\begin{tabular}{@{}c@{\hspace{0.4em}} l c@{\hspace{0.2em}} ccc c@{\hspace{0.2em}} ccc c@{\hspace{0.2em}} ccc c@{\hspace{0.2em}} c@{}}
\toprule
&& \phantom{} & \multicolumn{3}{c}{$d=1$} &  \phantom{} & \multicolumn{3}{c}{$d=2$} &  \phantom{} & \multicolumn{3}{c}{$d=3$} & \phantom{} \\
\cmidrule{4-6} \cmidrule{8-10} \cmidrule{12-14} 
& Model && R$@1K$ & R$@5K$  & R$@10K$ && R$@1K$ & R$@5K$  & R$@10K$  && R$@1K$ & R$@5K$  & R$@10K$  \\
\midrule
&\textbf{Learning to Cut (Full)} && 8.18 & 24.44 & 30.40 && 15.30 & 48.26 & 59.50 && 19.18 & 64.30 & 79.42 \\
\midrule
&w/o audio && 6.30 & 22.65 & 31.88 && 12.61 & 44.56 & 61.85 && 16.54 & 59.37 & 82.13 \\
&w/o auxiliary  && 4.91 & 20.64 & 23.23 && 10.08 & 43.95 & 48.85 && 13.78 & 61.29 & 67.95 \\

\bottomrule
\end{tabular}
\caption{\textbf{Ablation study.}  We evaluate our method against its variants: without the visual stream, without the audio stream, and without the auxiliary task. We observe that our design choices for modeling audiovisual information and jointly training learning to cut with a single-sided cut probability provides the edge to achieve better ranking results. } 
\label{table:ablation-complete}
\end{table*}

\noindent\textbf{Implementation Details.} We first extract frames for each scene video at $24$ fps. In terms of backbones, we use ResNexT-101-3D \cite{hara3dcnns} pre-trained on Kinetics \cite{kay2017kinetics} for the visual stream and ResNet-18 pre-trained on VGGSound \cite{Chen20} for the audio stream. We freeze each of them and extract features from each of the last convolutional layers after Global Average Pooling. We extract snippet features in a sliding window strategy with window size of $T=16$ frames and stride of $\Delta=8$. The feature dimensions are 2048 for the visual backbone and 512 for the audio backbone. We concatenate these features into a $2560$-dimensional feature vector. The final output is the feature map $F_n \in R^{t_{n} \times 2560}$. We jointly train the contrastive learning module (CLM) and the auxiliary task module (ATM) with an initial learning rate of $0.003$ using Adam optimizer. We use  $l_1=2$ layers with a reduction factor of $c=2$ for the CLM, and $l_2=1$ layers for the ATM. We reduce the learning rate by a factor of $0.9$ if the validation loss does not decrease after an epoch. We choose $\lambda_1=1$ and $\lambda_2=0.2$ as trade off coefficients in Equation (\ref{equation:loss}), such that both losses are of the same scale. While we train both tasks together, we perform inference through a two-stage prediction by first choosing the top $M=30\%$ scoring snippets from the ATM and then having CLM rank these retrieved pairs. \\
\noindent\textbf{Baselines.} Our main task is to rank each of the pairs according to their similarity score. Ideally, the pairs that first appear in the ranked list are the ones that are a better fit to form the cut, \ie the right places to cut. Below, we define all the methods that will be compared on this task. \\
\noindent\underline{Random Baseline}: We assign a uniform random score for each pair of clips in the validation subset.\\
\noindent\underline{Audio-visual baseline}: Since the ranking is given by a similarity score, we can use the raw backbone features directly (\ie without passing them through CLM and ATM) to measure their correlations as the ranking score. In our comparisons, we study three alternatives: (i) using only visual features (visual), (ii) using only audio features (audio), and (iii) concatenating audio-visual features (Audio-visual).\\
\noindent\underline{Learning to Cut (Single-stage)}: We use the CLM scores for all the pairs to perform the ranking, \ie ATM is not used here. We refer to this baseline as Ours (\textit{w/o multi-stage}).\\
\noindent\underline{Learning to Cut (Full):} As explained earlier, we first use the individual clip scores from ATM to choose the top $M$ scoring clips, which are then ranked using CLM similarity scores. \\
\textbf{Inference time.} At run time, features are pre-computed and cached and the feature similarity (cut ranking) runs on the fly. Extracting features takes about $6$ seconds per minute of video ($240$ fps). Computing similarity can be done efficiently via dot product, which computation time is negligible compared to the forward pass.

\subsection{Ranking Results}

\noindent\textbf{Comparison with baselines}
We compare our Learning to Cut method against the different baselines in Table \ref{table:baselines_comparison}. We report Recall $R@\alpha K$ with $\alpha\in\{1,5,10\}$ and under distances $d\in\{1,2,3\}$ clips (or $d\in\{0.33, 0.67, 1\}$ seconds) from the ground truth cut. We observe how challenging the task is by looking at the Random baseline performance. In the most strict setting ($R@1K$ and $d=1$), this method can only retrieve $0.6\%$ of the positive pairs. Even in the loosest setting ($R@10K$ and $d=3$), it only retrieves $\sim34\%$ of the positive pairs. Since sharp visual changes often occur between shots in continual editing, the similarity score of cross-shot clips is very low \cite{liu2020multi}, even at the ground truth cuts. Thus, the raw visual features perform poorly as well. 

When combining both raw audio-visual features, we observe a different trend. Even though the results are low, they are better than random chance. We attribute this performance discrepancy (visual \textit{vs.} audio-visual raw features) to the fact that most audios are continuous across the given shot pairs; making the audio features to spot temporal discontinuities. However, that is not enough. Overall, we observe that raw visual features are not appropriate enough to offer a good margin in ranking performance with respect to random chance. 
In contrast, both variants of our models outperform random chance (and the baselines) by large margins across all metrics. Indeed, the most important finding is the effectiveness of the two-stage inference, which improves the performance of the single-stage model by two or even three times in each of the metrics. This finding shows the importance of combining CLM and ATM, which mimics the process of first finding individual cut moments and then finding pairs that make a smooth transition across shots. We find that there is no significant difference when evaluating only on scenes from movies that were not seen at training.

\begin{figure*}[ht!]
  \centering
  \includegraphics[width=0.9\linewidth]{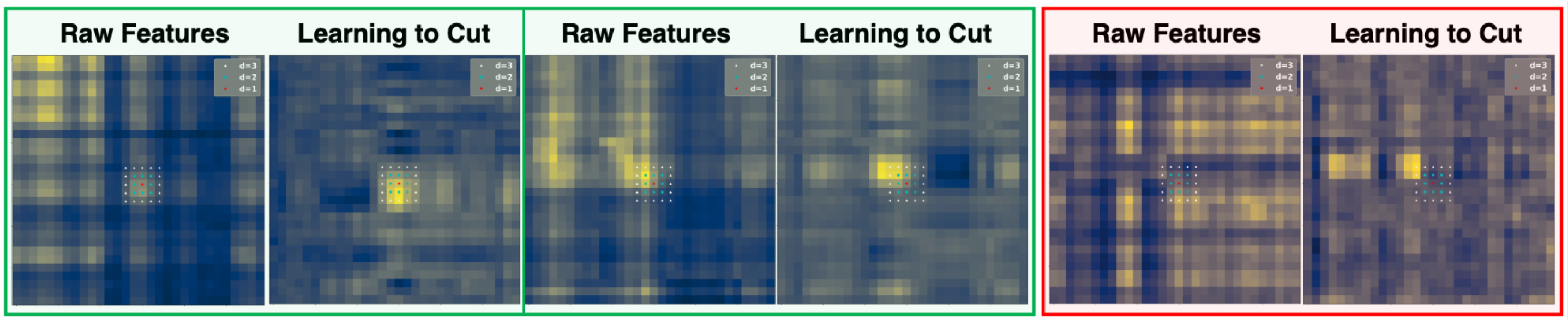}  
  \caption{\textbf{Qualitative Results}. We show the pair-wise feature similarity across two consecutive shots before and after learning to cut. The lighter the color the higher the correlation. Sharper regions around the center of the matrices corresponds to cases where Learning to Cut is able to retrieve the correct ground truth (green box). We observe that Learning to Cut is able to make the cross-shot similarity spike in short temporal regions. In contrary, raw features' cross-shot similarity activates sparsely.}
\label{fig:qualitative}
\end{figure*}
\begin{figure}[h!!]
\begin{subfigure}{\linewidth}
\centering
  \includegraphics[width=0.9\linewidth]{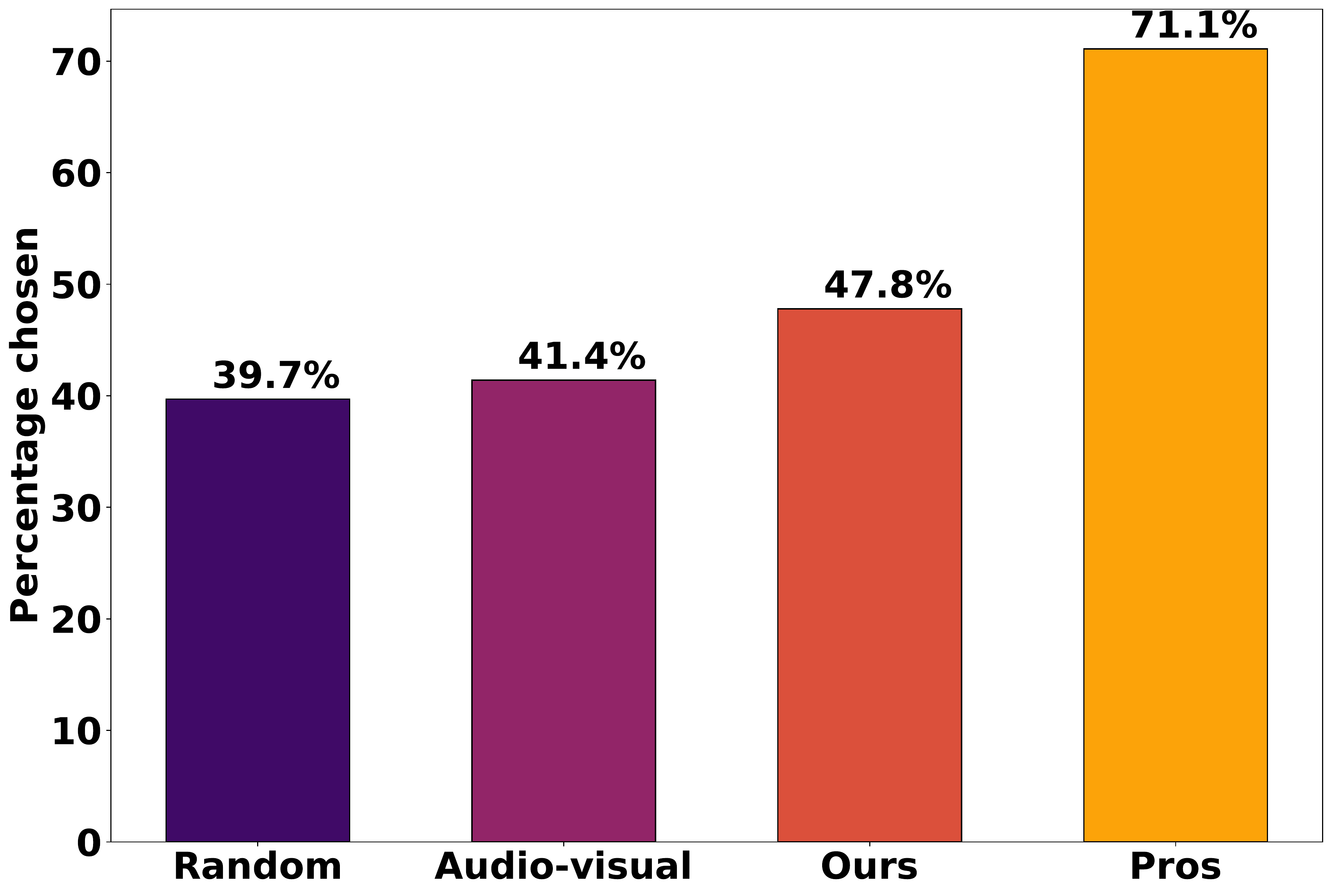}
  \caption{\textit{One vs All} percentage of times each source of cut is preferred.}\label{fig:absolute-human-study}
\end{subfigure}
\centering
\begin{subfigure}{\linewidth}
\centering
  \includegraphics[width=0.9\linewidth]{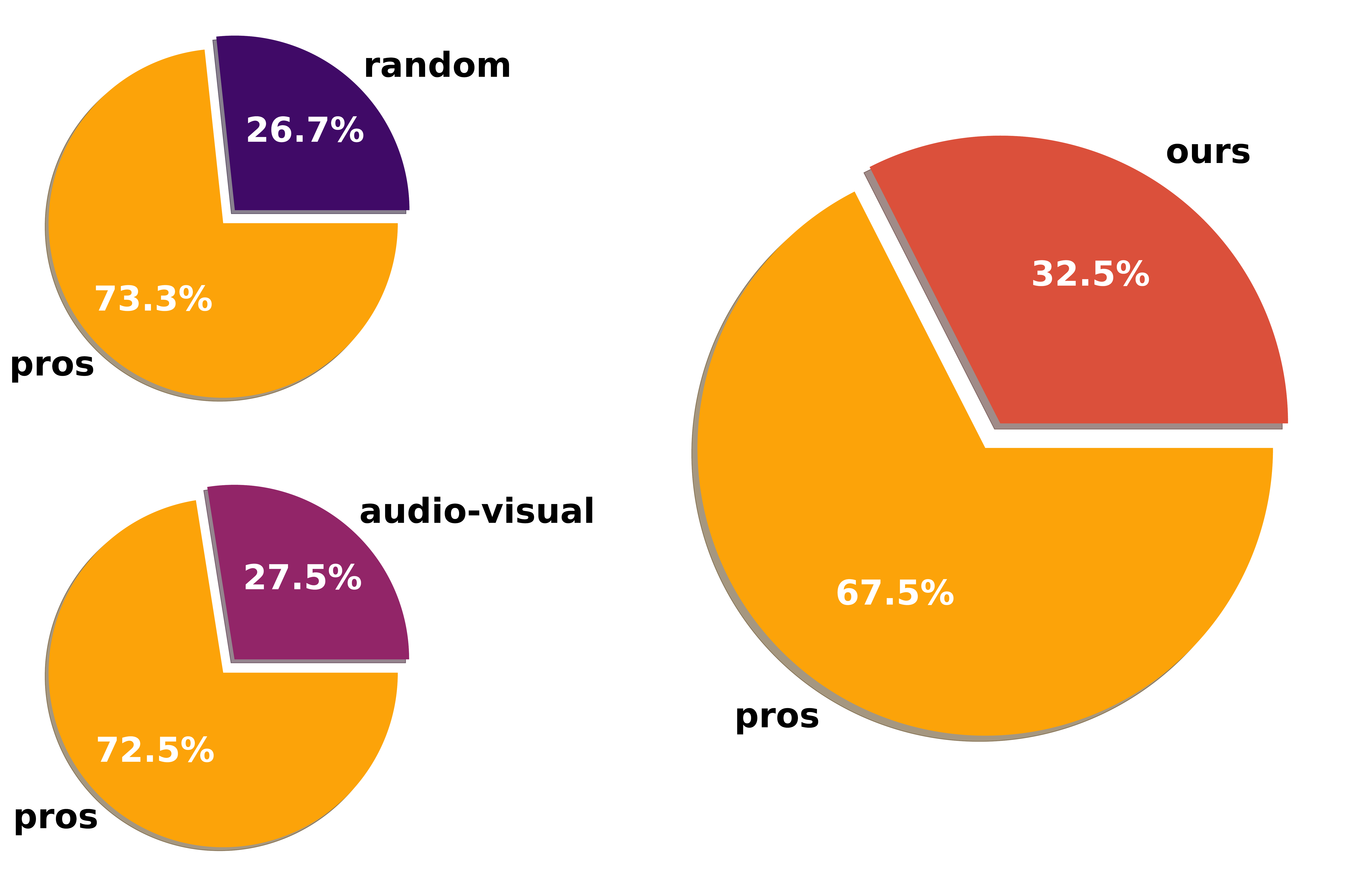}
  \caption{\textit{Pair-wise} percentage of times each source of cut is preferred over the professional cuts.}\label{fig:relative-human-study}
\end{subfigure}

\caption{\textbf{Human study.} We show the absolute percentages per cut source \ref{fig:absolute-human-study}, and the chosen percentage of each of them relative to professional cuts \ref{fig:relative-human-study}. These results provide further evidence that our approach offers better rankings than the Audio-visual and random baselines.
}
\label{fig:human-study}
\end{figure}

\noindent\textbf{Ablation Study.}
To validate the design choices in our Learning to Cut model, we remove each of its main components, evaluate the resulting variant model, and report all ablation results in  Table \ref{table:ablation-complete}. 
We observe that modeling both Audio and Visual information jointly provides and edge in performance. This trend is particularly evidenced for the stricter ranking metrics (\eg $R@1K$). We attribute these results to the fact that a good cut is made of both: visual compositions and sound; and having a single modality is not enough to do a good ranking. Finally, we observe that one key component of our model is the auxiliary task. When it is not used to facilitate the learning process (\ie when $\lambda_2=0$), the model's performance drops several points in each metric. We validate the auxiliary task is an essential component of our model. This task aligns with the editing process, whereby the editor picks a place where to cut, and only then, they pick the best match to glue together \cite{filmgrammar, berthouzoz2012tools}. More detailed ablations can be found in the \textbf{supplementary material.}

\subsection{Human Studies}
We use our method for ranking cut plausibility on raw video footage licensed from Editstock.com. We licensed five different projects including more than eight hours of raw/unedited footage. Unlike previous experiments, we now address the cut ranking problem in the same setting an editor would encounter it. In this case, the method takes as input two untrimmed unedited videos. The task is to produce potential cut locations in each of these videos. While we have shown that our method offers good ranking results in our proxy task, does it work for practical scenarios, where videos are unedited and untrimmed? 

To answer this question, we conduct a human study with ten professional video editors. Although we could have invited general audience to our study, we decide not to do so; it has been shown \cite{smith2008edit} that the general population is unable to perceive the fine-grained details of good cuts. We design a study that asks participants to select the best cut among two options, which can be either taken from the professional editors' cuts \footnote{We invited professional editors to create video cuts from five Editstock projects. From about eighth hours of footage, the editors curated $120$ cuts.} generated by the random method, generated by the audio-visual baseline, or generated by our proposed approach. We conduct the user study in a one vs. one manner: \ie we mixed all possible combination of choices for the participants, and all the options were faced against each other. Thus, every alternative appeared the same number of times. We report the results in Figure \ref{fig:human-study}. Interestingly, our method's cuts are chosen by the participants more often than those from the baselines. However, the editors can clearly identify the professionally curated cuts. Despite the progress made by this work, there is still a long way to generate cuts with professional quality. We do believe, though, that our approach and baselines are first steps towards this direction.

\noindent\textbf{Qualitative Results} We show a pair-wise feature correlation between a pair of adjacent shot in figure \ref{fig:qualitative}. Each entry of the matrix is the similarity between a snippet of the left and right shot of the cut. The axes are center around the cut, such that the center of the matrix has the similarities of the positive pairs. We observe how our method transforms the features and make them spike around the cut's region.

\section{Conclusion}
\label{sec:conclusion}
We introduced the task of cut plausibility ranking for computational video editing. We proposed a proxy task that aligns with the actual video editing process by leveraging knowledge from already edited scenes. Additionally, we collected more than 260K edited video clips. Using this edited footage, we created the first method capable of ranking cuts automatically, which learns in a data-driven fashion. We benchmarked our method with a set of proposed metrics that reflect the model's level of precision at retrieval and expertise at providing tighter cuts. Finally, we used our method in a real-case scenario, where our model ranked cuts from non-edited videos. We conducted a user study in which editors picked our model's cuts more often compared to those made by the baselines. Yet, there is still a long way to match editors' expertise in selecting the most smooth cuts. This work aims at opening the door for data-driven computational video editing to the research community. Future directions include the use of fine-grained features to learn more subtle patterns that approximate better the fine-grained process of cutting video. Additionally, other modalities such as speech and language could bring benefits for ranking video cuts.  

\noindent\textbf{Acknowledgments} This work was supported by the King Abdullah University of Science and Technology (KAUST) Office of Sponsored Research through the Visual Computing Center (VCC) funding.

{\small
\bibliographystyle{ieee_fullname}
\bibliography{egbib.bib}
}
\clearpage
\appendix
\section*{Supplementary Material}

\section{Learning to Cut by Users- Toy Example}

Please visit: \textit{\textbf{\url{https://alejandropardo.net/publication/learning-to-cut/}}} for code and complete supplementary material. 

To illustrate to the reader a toy example of Learning to Cut we included a folder called \href{./spot-the-real-cut.zip}{Spot-the-Real-Cut}. We encourage the reader to open the \textit{html} file contained in the supplementary material folder and try to choose the more suitable cuts. You will have to wait around 15 seconds for the link to load all the videos. There are going to be 30 examples, each of them showing a pair of cuts. One of them breaks continuity, while the other is an actual cut made by a professional editor. The task is simple: \textbf{choose the cut that is real}. To play the video, click on top of it. To decide what you consider is the real cut, click on the button "This cut is real" below the clip. At the end of the study, you will see what percentage of cuts from the ones chosen were actually real. The purpose of this toy example is to illustrate that there is a signal that a model could learn to Learn how to cut. Such a signal is the one that Learning to Cut is aiming to leverage.

\section{DatasetStatistics}

\begin{figure}[h!]
  \centering
  \includegraphics[width=\linewidth]{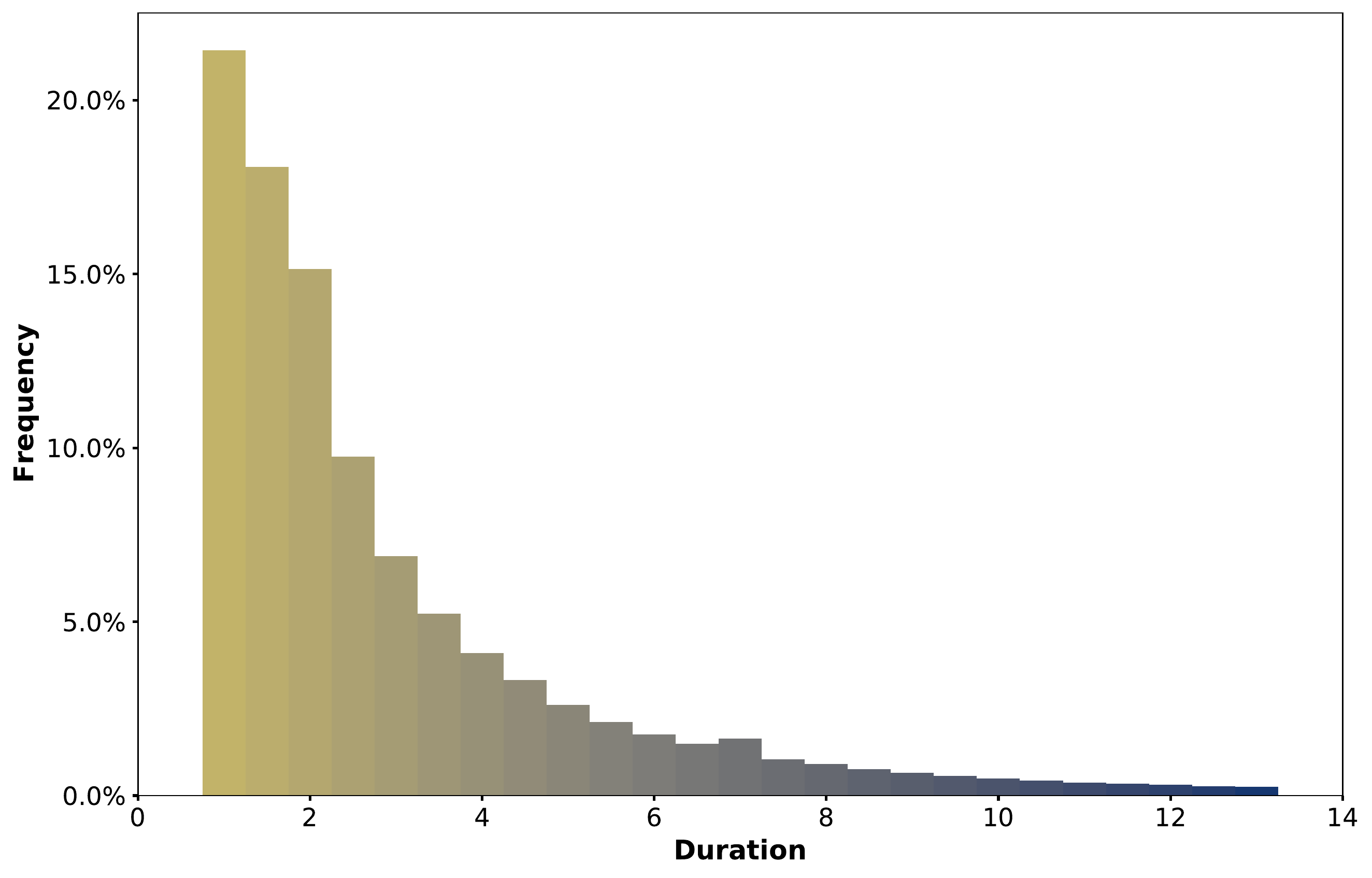} 
  \caption{\textbf{Shots' duration distribution.}}
\label{fig:shots-duration}
\end{figure}

Additional statistics are shown below. Figure \ref{fig:shots-genre} shows the distribution of number of shots per genre along the dataset. Figure \ref{fig:shots-duration} shows the shots-duration's distribution. Most of the shots in the movies are shorter than 2 seconds. This challenging property comes from the fast-pace edits of action scenes, where the shot duration is tipically short. \\
\vspace{-10pt}

\begin{figure}[h!]
  \centering
  \includegraphics[width=\linewidth]{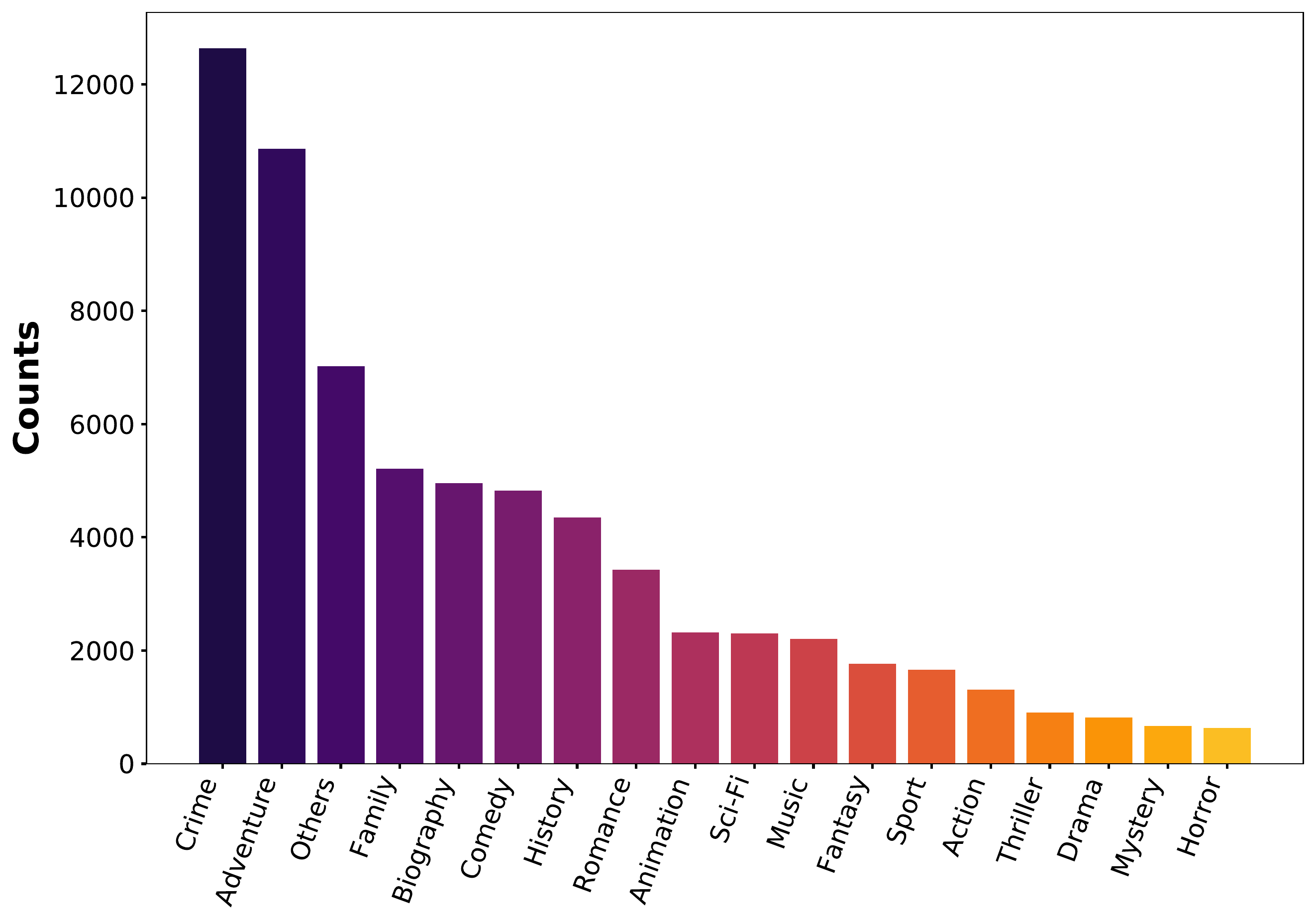} 
  \caption{\textbf{Distribution of shots per genre.}}
\label{fig:shots-genre}
\end{figure}


\section{Generalization to unedited set.}

\noindent\textbf{Qualitative results.} In table \ref{table:generalization}, we report the qualitative number of our method on the unedited set. We see that the real task is really challenging as the number drop significantly from the proxy task. However, we observe the same trend in the results, our method outperforms the baselines. This results show how challenging is the tasks in a real-world scenario. Thus, future methods have to put effort to solve first the proxy as this results will be reflected in the real task.

\begin{table*}[!ht]
\footnotesize
\small
\centering
\begin{tabular}{@{}c@{\hspace{0.4em}} l c@{\hspace{0.2em}} ccc c@{\hspace{0.2em}} ccc c@{\hspace{0.2em}} ccc c@{\hspace{0.2em}} c@{}}
\toprule
&& \phantom{} & \multicolumn{3}{c}{$d=1$} &  \phantom{} & \multicolumn{3}{c}{$d=2$} &  \phantom{} & \multicolumn{3}{c}{$d=3$} & \phantom{} \\
\cmidrule{4-6} \cmidrule{8-10} \cmidrule{12-14} 
& Model && R$@1K$ & R$@5K$  & R$@10K$ && R$@1K$ & R$@5K$  & R$@10K$  && R$@1K$ & R$@5K$  & R$@10K$  \\
\midrule
&Random && 0.83 & 3.33 & 3.33 && 1.67 & 8.33 & 15.00 && 3.33 & 17.50 & 30.00\\
&Audio-visual && 1.67 & 4.17 & 5.00 && 2.50 & 5.83 & 10.83 && 2.50 & 11.67 & 20.00\\
&\textsc{\textbf{Ours}} && 0.83 & 2.50 & 8.33 && 1.67 & 10.83 & 30.83 && 5.83 & 17.50 & 34.67 \\
\bottomrule
\end{tabular}
\caption{\textbf{Generalization to unedited videos.} We showcase our model's results in a real-case scenario where it processes raw unedited footage. We report the same metrics as before by comparing our results with the cuts of professional editors.} 
\label{table:generalization}
\end{table*}

\section{Additional Ablation Study}
\noindent\textbf{Ablation Study.} In Table \ref{table:ablation-complete}, we report the complete numbers for the ablation study. Compared with Table 2 of the main manuscript, we show here $R@10K$ for every $d$. \\

\begin{table*}[!ht]
\centering
\footnotesize
\begin{tabular}{@{}c@{\hspace{0.4em}} l c@{\hspace{0.2em}} ccc c@{\hspace{0.2em}} ccc c@{\hspace{0.2em}} ccc c@{\hspace{0.2em}} c@{}}
\toprule
&& \phantom{} & \multicolumn{3}{c}{$d=1$} &  \phantom{} & \multicolumn{3}{c}{$d=2$} &  \phantom{} & \multicolumn{3}{c}{$d=3$} & \phantom{} \\
\cmidrule{4-6} \cmidrule{8-10} \cmidrule{12-14} 
& Model && R$@1K$ & R$@5K$  & R$@10K$ && R$@1K$ & R$@5K$  & R$@10K$  && R$@1K$ & R$@5K$  & R$@10K$  \\
\midrule
&\textsc{\textbf{Ours}} && 8.18 & 24.44 & 30.59 && 15.30 & 48.26 & 59.83 && 19.18 & 64.30 & 79.87 \\
\midrule
&w/o visual && 7.82 & 24.65 & 32.82 && 14.99 & 48.21 & 63.56 && 18.96 & 64.07 & 84.13 \\
&w/o audio && 6.30 & 22.65 & 31.88 && 12.61 & 44.56 & 61.85 && 16.54 & 59.37 & 82.13 \\
&w/o auxiliary  && 4.91 & 20.64 & 23.23 && 10.08 & 43.95 & 48.85 && 13.78 & 61.29 & 67.95 \\

\bottomrule
\end{tabular}
\caption{\textbf{Ablation study.}  We evaluate our method against its variants: without the visual stream, without the audio stream, and without the auxiliary task.} 
\label{table:ablation-complete}
\end{table*}

\begin{figure*}[h!]
  \centering
  \includegraphics[width=0.9\linewidth]{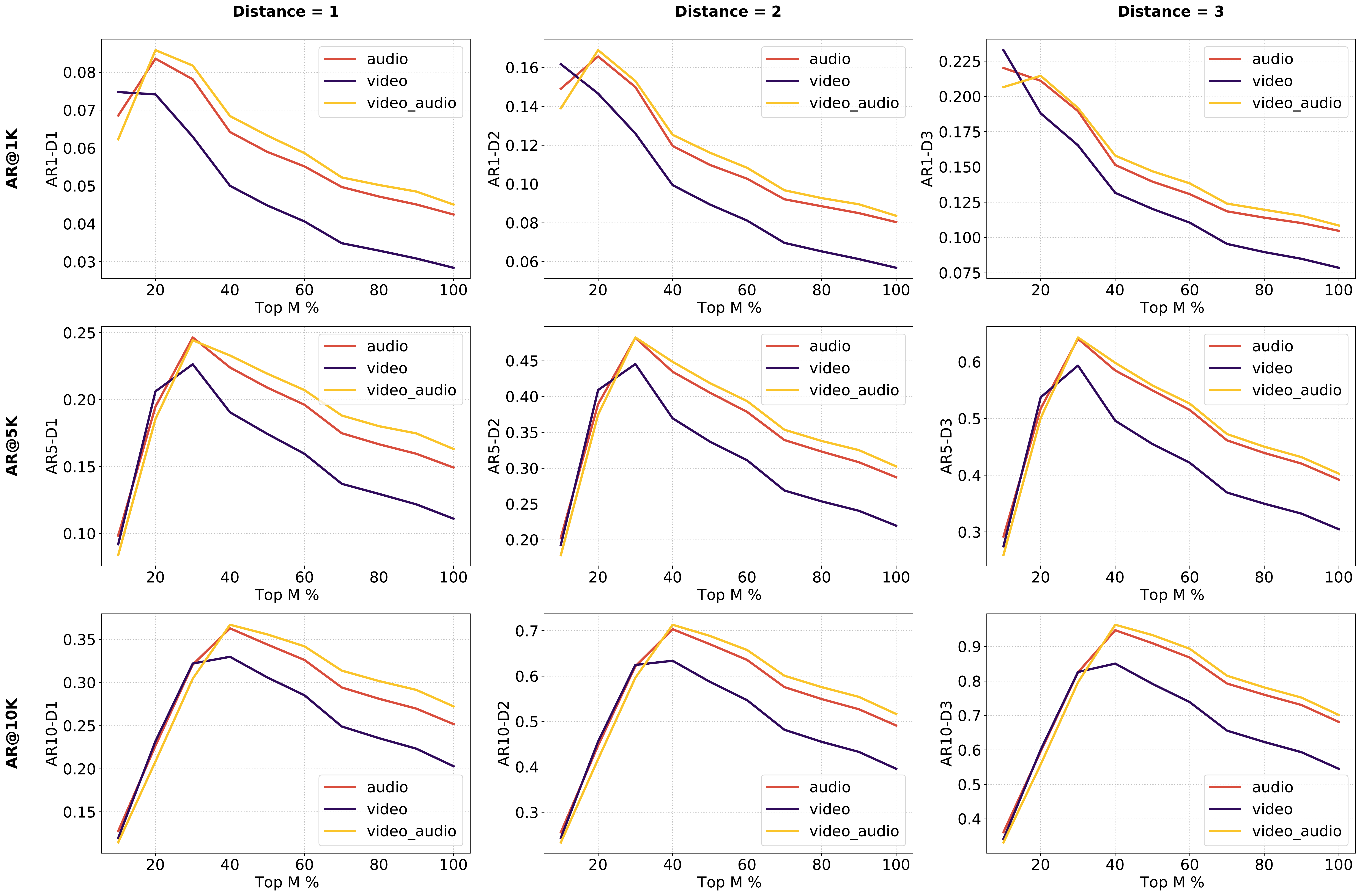} 
  \caption{\textbf{Influence of top M\% per metric.}}
\label{fig:topm}
\end{figure*}

\noindent\textbf{Impact of M.} Figure \ref{fig:topm} shows the impact of the top $M\%$ parameter in the two-stage prediction. We can see different peaks according to the metric that we are looking at; however, there is a clear pattern top $20\%$ favors the best $R@1K$, top $30\%$ $R@5K$, and top $40\%$ $R@10K$, no matter the distance. For the Unedited set we chose top $30\%$, since we were using the top-5 predictions. \\
\vspace{-10pt}

\section{Qualitative Results}

\begin{figure}[h!]
  \centering
    \begin{subfigure}{\linewidth}
      \includegraphics[width=\linewidth]{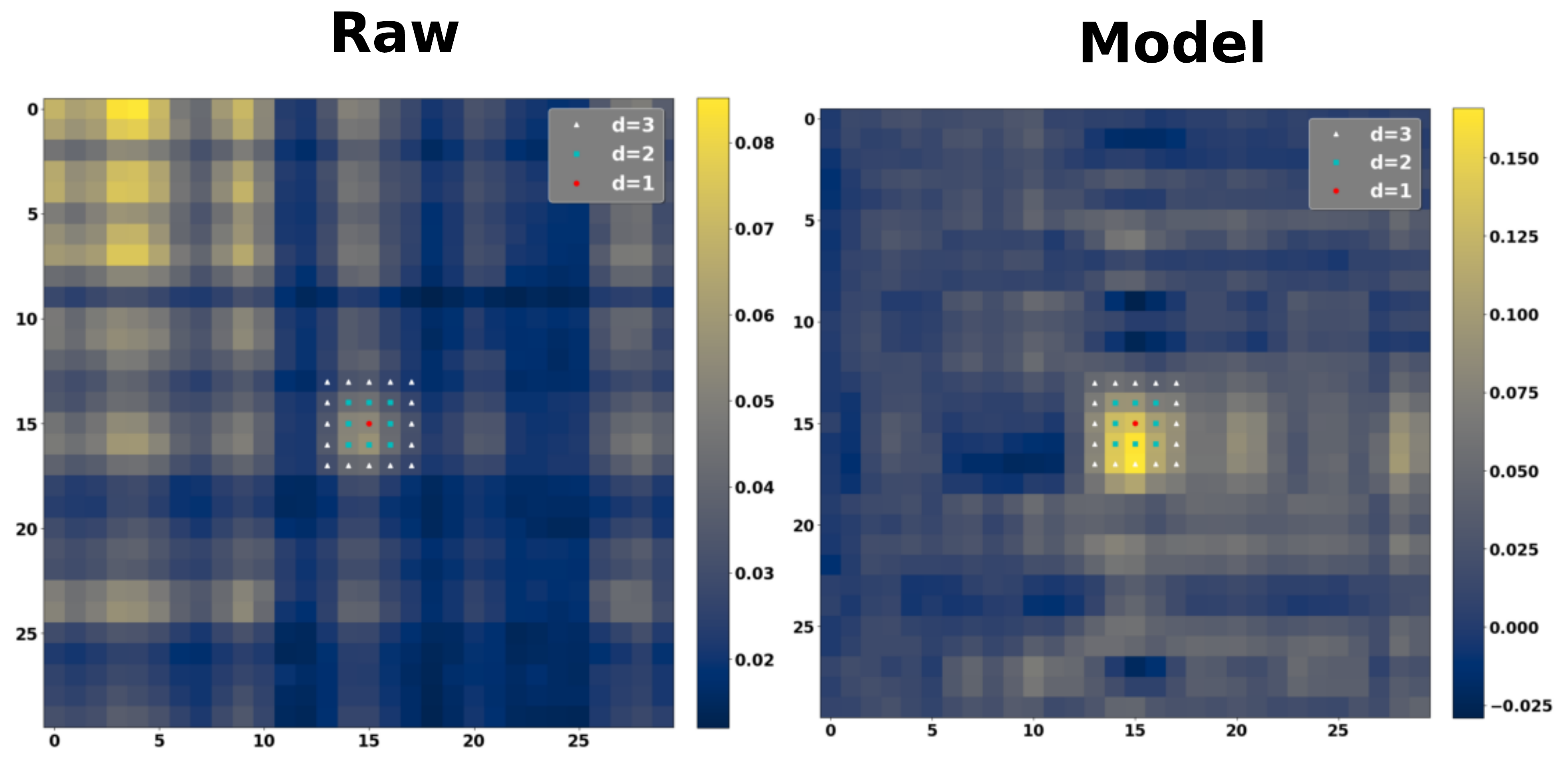}
     \caption{Correct prediction - Example 1}\label{fig:pos-ex1}
    \end{subfigure}
    \begin{subfigure}{\linewidth}
       \includegraphics[width=\linewidth]{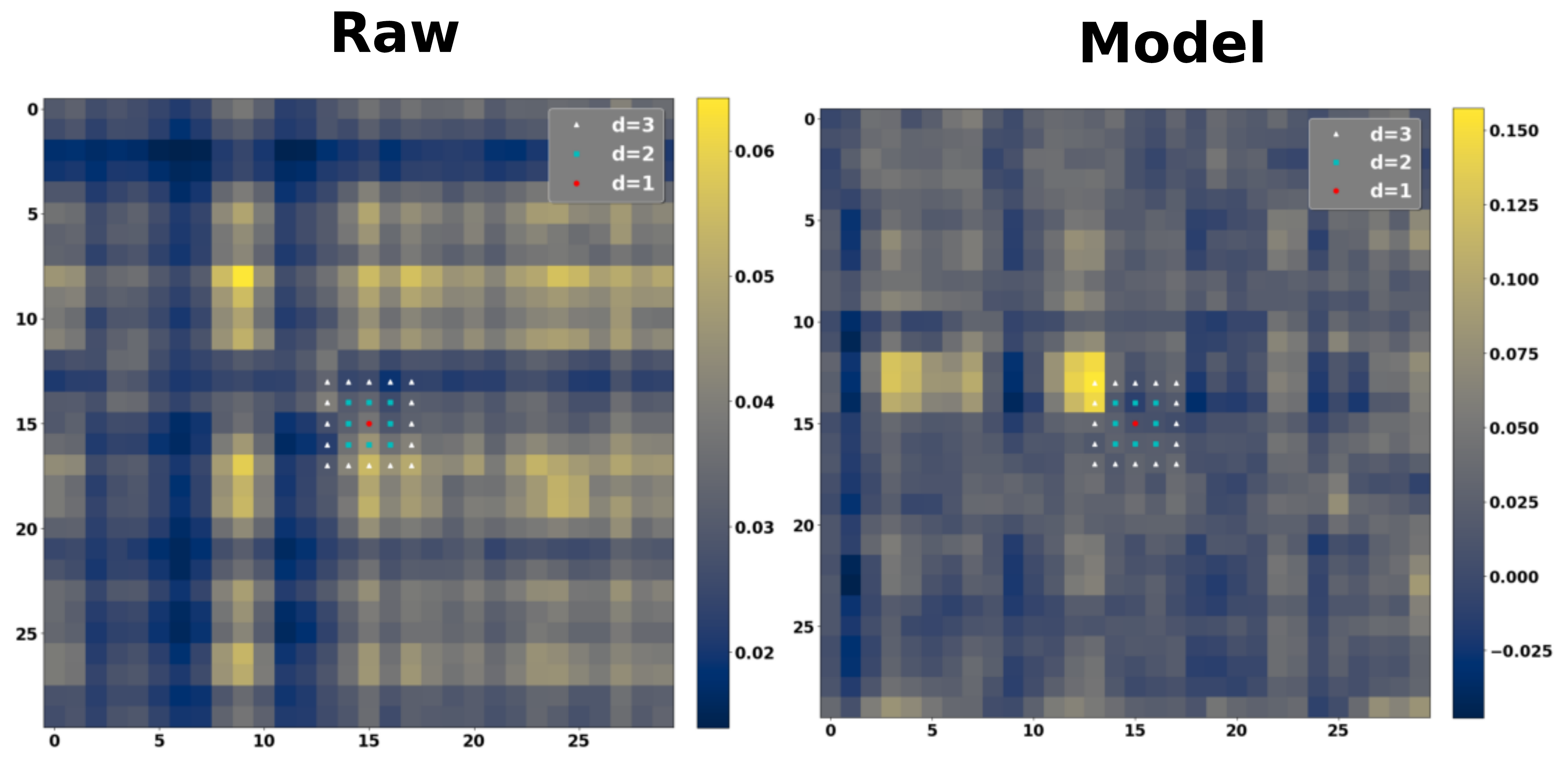}
     \caption{Correct prediction - Example 2}\label{fig:pos-ex2}
    \end{subfigure}
    \begin{subfigure}{\linewidth}
      \includegraphics[width=\linewidth]{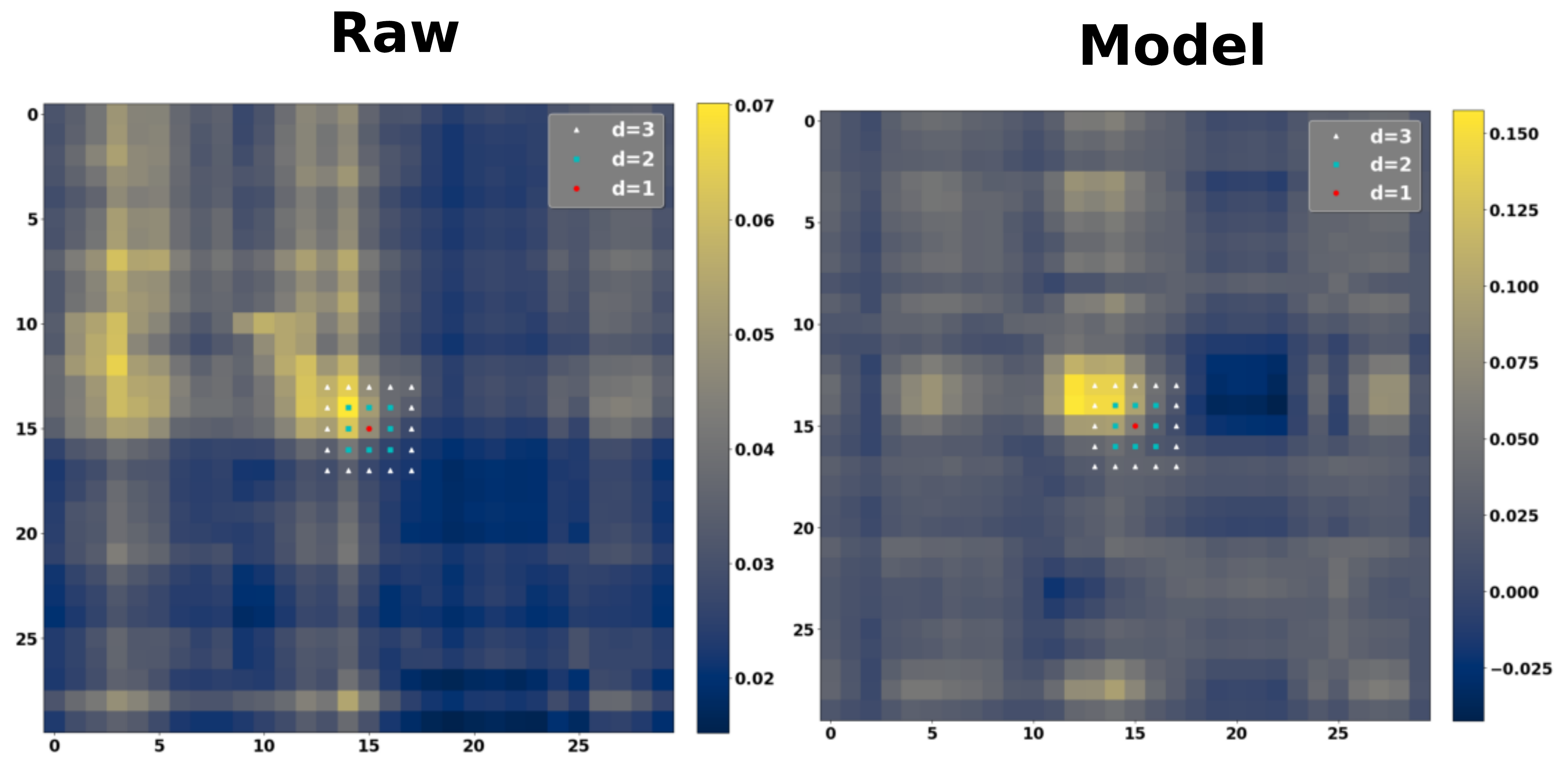}
     \caption{Correct prediction - Example 3}\label{fig:pos-ex3}
    \end{subfigure}
    
  \caption{\textbf{Feature similarities before and after Learning to Cut.} Examples where the feature similarities were moved correctly to the ground-truth region. Please zoom in for a better view.}
  \label{fig:qualitative-pos}
\end{figure}

\begin{figure}[h!]
  \centering
    \begin{subfigure}{\linewidth}
      \includegraphics[width=\linewidth]{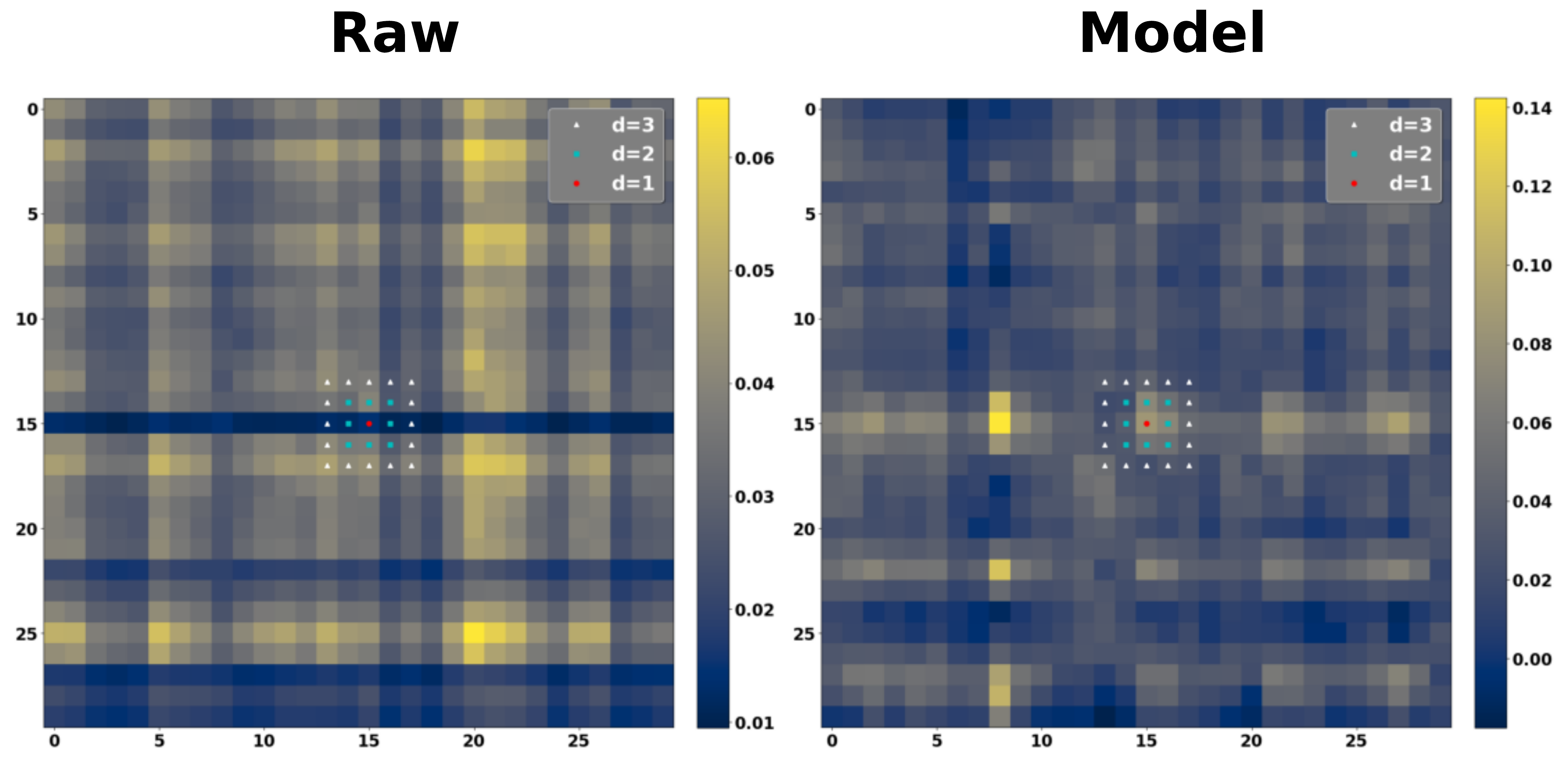}
     \caption{Incorrect prediction - Example 1}\label{fig:neg-ex1}
    \end{subfigure}
    \begin{subfigure}{\linewidth}
      \includegraphics[width=\linewidth]{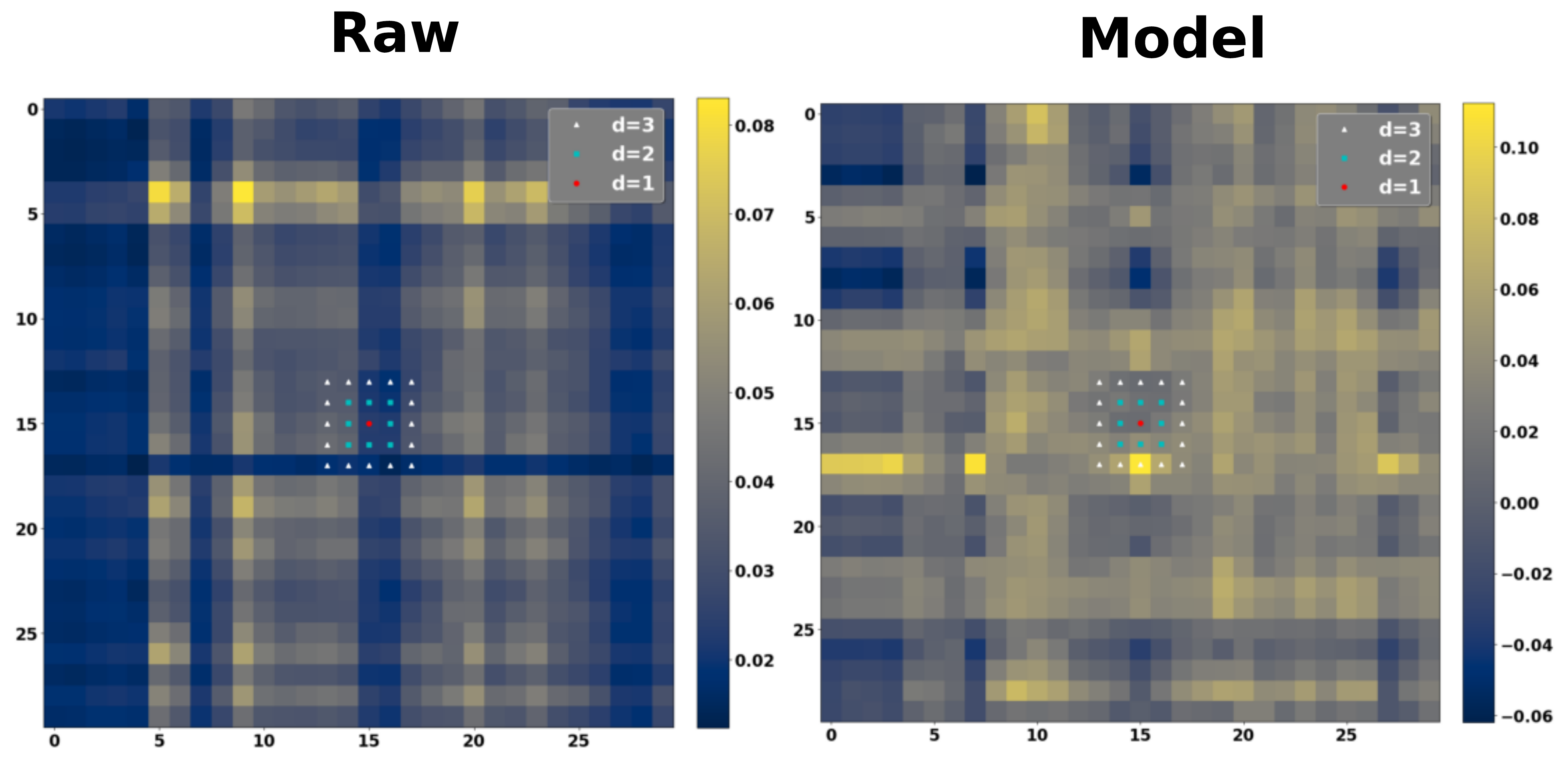}
     \caption{Incorrect prediction - Example 2}\label{fig:neg-ex2}
    \end{subfigure}
    \begin{subfigure}{\linewidth}
     \includegraphics[width=\linewidth]{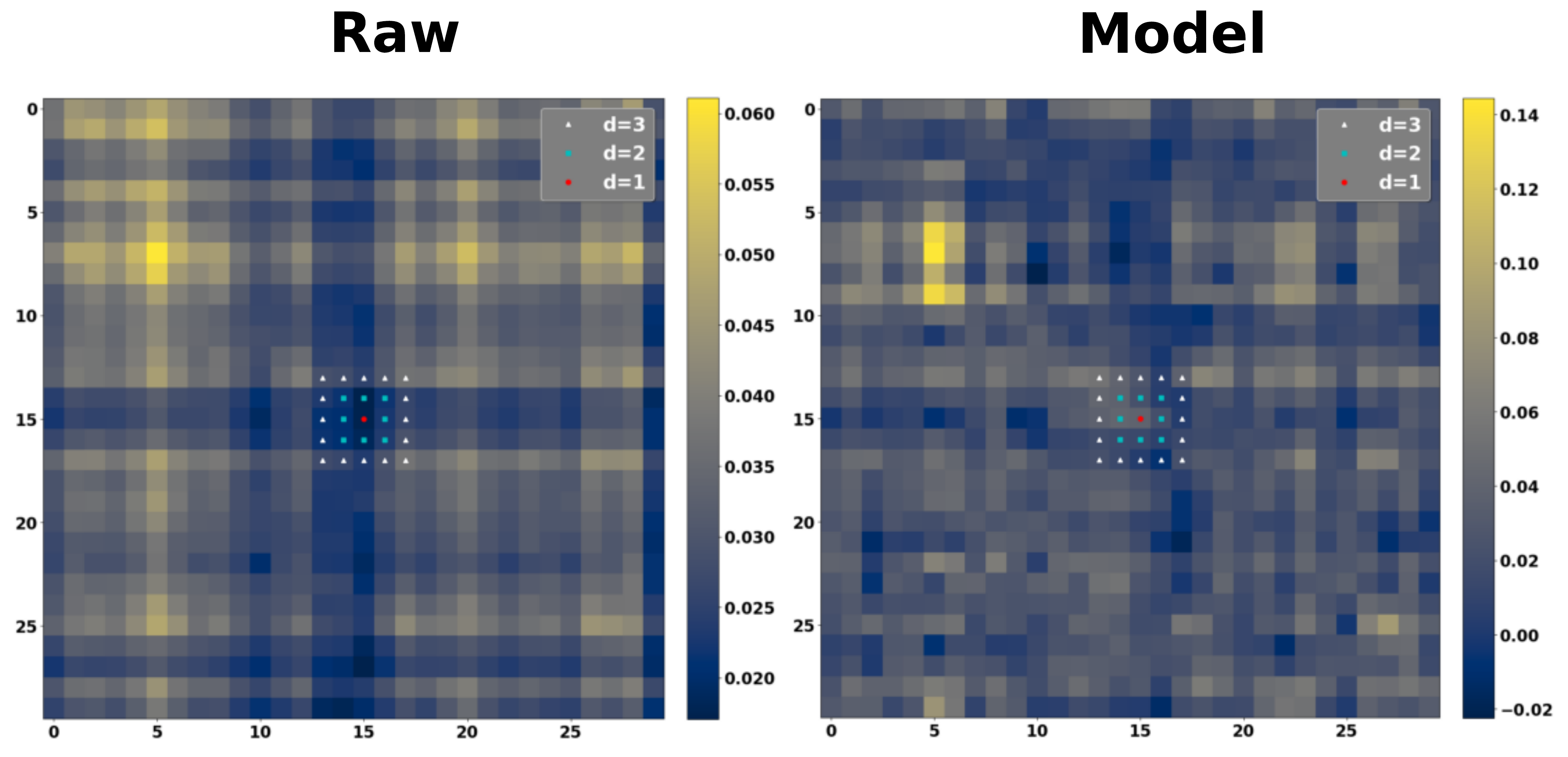}
     \caption{Incorrect prediction - Example 3}\label{fig:neg-ex3}
    \end{subfigure}
    
  \caption{\textbf{Feature similarities before and after Learning to Cut.} Examples where the feature similarities were moved out of the the ground-truth region. Please zoom in for a better view.}
  \label{fig:qualitative-neg}
\end{figure}

In Figure \ref{fig:qualitative-pos} and Figure \ref{fig:qualitative-neg} we show the feature temporal similarity of two candidate videos to be stitched together. The columns of the matrix represent snippets of video one, and the rows represent snippets of video 2. We show the similarity between these two set of snippets before (Raw) and after (Model) Learning to Cut. In this case all the cuts are in $(15, 15)$, we show in red the region of the ground-truth with distance $d=1$, in cyan the edge of the region for ground-truth with $d=2$, and in white the edge of the region for ground-truth with $d=3$. 
On the one hand, we observe in figure \ref{fig:qualitative-pos} that our model tends to localize the similarities around the actual cutting place and the ground-truth regions. In Figure \ref{fig:pos-ex1}, the most salient region is overlapping the ground-truth region after our model was applied, before it, the similarity spikes where located in a complete different place. Thus, our model was able to transform the video features such that the similarities spike around the cutting points. We observe similar behavior for Figure \ref{fig:pos-ex2} and Figure \ref{fig:pos-ex3}; however, the center of the spike region is a couple of spaces off the ground-truth region.
On the other hand, we can observe in Figure \ref{fig:qualitative-neg} some examples in which the spike of the similarities do not match the ground-truth region. Interestingly Figure \ref{fig:neg-ex1} and Figure \ref{fig:neg-ex2} show that the cutting point for one of the videos was predicted correctly (the spike happens along the 15th row); yet, the model was not able to find a cutting point for the second video that would match the ground-truth. This does not necessarily mean that the cutting point found the model is not correct. It means that did not match the cut made by the professional. The Figure \ref{fig:neg-ex3} shows a spike on a region that does not correspond with the ground truth. In this case, the model was not able to move the features away from the initial state, since the features were already spiking in a similar region before the model (raw column).
Regardless of the ground-truth region, Figure \ref{fig:qualitative-pos} and Figure \ref{fig:qualitative-neg} show that our model helps to sharpen feature similarities in specific regions across a pair of videos. The similarity spike is not as blur anymore as it was in the original features (Raw).
\textbf{Additional qualitative results with the actual clips ranking can be found on the attached files and slides. In the examples' files (\href{./Qualitative.zip}{Qualitative.zip}) the videos are name after their ranking, the real cut is also included.}

}

\end{document}